\documentclass{article}

\usepackage{microtype}
\usepackage{graphicx}
\usepackage{subfigure}
\usepackage{booktabs} 

\usepackage{xcolor}

\usepackage{hyperref}

\usepackage[accepted]{icml2020}

\icmltitlerunning{Neural Clustering Processes}

\usepackage{amsfonts}       
\usepackage{mathtools}
\usepackage{enumitem}

\usepackage{amsmath}
\usepackage{amsthm}
\usepackage{amssymb}
\usepackage{pifont}
\newcommand{\xmark}{\ding{55}}%

\newcommand{\eq}{\begin{equation*}}
\newcommand{\en}{\end{equation*}}
\newcommand{\eqa}{\begin{eqnarray*}}
	\newcommand{\ena}{\end{eqnarray*}}
\newcommand{\eqn}{\begin{equation}}
\newcommand{\enn}{\end{equation}}
\newcommand{\be}{\begin{equation}}
\newcommand{\ee}{\end{equation}}
\newcommand{\eqan}{\begin{eqnarray}}
\newcommand{\enan}{\end{eqnarray}}

\newcommand{\nn}{\nonumber}

\newcommand{\s}{ {\bf s} }
\newcommand{\cc}{ {\bf c} }
\newcommand{\bfa}{ {\bf a} }
\newcommand{\bb}{ {\bf b} }

\newcommand{\ba}{\textbf{a}}

\newcommand{\y}{ {\bf y} }
\newcommand{\x}{ {\bf x} }
\newcommand{\z}{ {\bf z} }

\usepackage[noabbrev,capitalize,nameinlink]{cleveref}

\usepackage{amsfonts}

\newcommand{\pmat}{\begin{pmatrix}}
\newcommand{\pman}{\end{pmatrix}}

\graphicspath{{./figures/} }

\usepackage{authblk}

\begin{document}

\twocolumn[
\icmltitle{Neural Clustering Processes}





\begin{icmlauthorlist}
\icmlauthor{Ari Pakman}{to}
\icmlauthor{Yueqi Wang}{to,yw}
\icmlauthor{Catalin Mitelut}{to}
\icmlauthor{JinHyung Lee}{to}
\icmlauthor{Liam Paninski}{to}
\end{icmlauthorlist}

\icmlaffiliation{to}{Columbia University}
\icmlaffiliation{yw}{Now at Google}
\icmlcorrespondingauthor{Ari Pakman}{aripakman@gmail.com}

\icmlkeywords{Machine Learning}

\vskip 0.3in
]



\printAffiliationsAndNotice{}  

\begin{abstract}
Probabilistic clustering models (or equivalently, mixture models) are
basic building blocks  in countless statistical models and involve latent
random variables over discrete spaces.
For these models, posterior inference methods can be inaccurate and/or very slow.  In this work we introduce deep network architectures trained with labeled samples from any generative model of  clustered datasets. At test time, the networks generate approximate posterior samples of cluster labels for any new  dataset of arbitrary size.
We develop two complementary approaches to this task, 
requiring  either O(N) or O(K)  network forward passes per dataset,  
where N is the dataset size and  K the number of clusters.
Unlike previous approaches, our methods sample  
the   labels of all the data points from a well-defined posterior,
and can learn nonparametric Bayesian posteriors
since they do not limit the number of mixture components.
As a scientific application, we present a novel approach to neural spike sorting for high-density multielectrode arrays. 
\end{abstract}




\section{Introduction}
Probabilistic clustering models (or equivalently, mixture models) are
a staple of statistical modelling
in which a discrete latent variable is introduced for each observation,
indicating its mixture component identity.
Popular inference methods in these models fall into two main classes. 
When exploring the full posterior is crucial (e.g. there is irreducible uncertainty about the latent structure or many separate local optima exist),  the method of choice is Markov Chain Monte Carlo (MCMC)~\citep{neal2000markov,jain2004split}. 
This method is asymptotically accurate but time-consuming, with convergence that is difficult to assess. 
Models whose likelihood and prior are non-conjugate
are particularly challenging, since in general in these cases the model parameters cannot be marginalized and
must be kept as part of the state of the Markov chain. 
Alternatively, variational methods~\citep{Blei:2004:VMD,kurihara2007collapsed,
hughes2015reliable} are typically much faster but do not come with accuracy guarantees. 

As a third alternative,
in recent years there has been steady progress on
amortized inference methods, and such is the spirit of
this work.
Concretely, 
we propose novel techniques  
to perform amortized approximate posterior inference 
over discrete latent variables in 
mixture models. 
We consider two possible 
product expansions of the mixture posteriors,
and in each expansion 
we use neural networks to express 
conditional factors 
in terms of fixed-dimensional, distributed representations that respect the permutation symmetries imposed by the discrete variables. A major advantage of our approach, compared 
to previous approaches to amortized clustering, 
is its ability to handle an arbitrary number of clusters
from a well defined posterior. 
This makes the methods a natural choice for 
nonparametric Bayesian models, 
such as Dirichlet process mixture models (DPMM), and their extensions. Moreover, the methods can be applied to both conjugate and non-conjugate models.

The term `amortization' refers 
to the process of investing computational resources to train 
a model that is later used for very fast posterior inference~\citep{gershman2014amortized}. 
Concretely, in a model with observations $x$
and latent variables $z$, the amortized approach learns a parametrized function $q_{\theta}(z|x)$ that approximates $p(z|x)$ for any $x$; learning the  parameters $\theta$ may be 
challenging, but once $\theta$ is in hand 
evaluating $q_{\theta}(z|x)$ for new data $x$ is fast. 

The amortized inference literature can be coarsely divided into two approaches.
On one side, the variational autoencoder (VAE) approach~\citep{kingma2013auto}, with roots in the 
wake-sleep algorithm~\citep{hinton1995wake}, 
learns $q_{\theta}(z|x)$ along with the generative model
$p_{\phi}(x|z)$. Here  $p(z)$ 
is usually a known simple distribution.


Our work corresponds to the alternative case: 
 a generative model
 $p(x,z)$ is postulated, and posterior inference is the main focus of
the learning phase. 
Amortized methods in this case usually involve 
a degree of specialization to the particular generative model
of interest. Examples include
methods developed 
for Bayesian networks~\citep{stuhlmuller2013learning}, 
sequential Monte Carlo~\citep{paige2016inference}, 
probabilistic programming~\citep{ritchie2016deep,le2016inference}, neural decoding~\citep{Parthasarathy}
and particle tracking~\citep{pmlr-v80-sun18b}.
Our work is specialized to the case 
where the latent variables are discrete 
and their range is not fixed beforehand.

After training a neural architecture using
labeled samples from a particular generative model, we can obtain independent, parallelizable, approximate posterior samples of the discrete variables for any new set of observations of arbitrary size, with no need for expensive MCMC steps.
These samples can be used 
(i) to approximate expectations,
(ii) as high quality importance samples,
or
(iii) as independent Metropolis-Hastings 
proposals. 

In~\cref{sec:generative} we introduce generative mixture models and 
present two distinct
expansions of the posterior distribution.
In~\cref{sec:NCP} and~\cref{sec:CCP} we present 
neural architectures to model the factors of each expansion,
along with their objective functions.
In \cref{sec:examples} we present two simple examples to illustrate the methods.
In~\cref{sec:related} we review related works. 
In~\cref{sec:QE} we discuss quantitative evaluations of the new  methods. 
We close in~\cref{sec:spike_sorting} with a neuroscientific application 
to spike sorting 
for high-density multielectrode probes.
The Supplementary Material (SM)
contains details on the architectures, the spike-sorting application, and an extension of these ideas to particle tracking.\footnote{An 
 early version appeared in~\cite{pakman2018amortized,wang2019spike}.
 Similar methods were applied to amortized permutations 
 in~\cite{pakman2019npp}.
 }

\section{Generative Mixture Models}
\label{sec:generative}
We start by presenting mixture models
from the 
perspective of probabilistic models for clustering~\citep{mclachlan1988mixture}.
The latter introduce random variables $c_i$ denoting  the cluster number to which the data point $x_i$ is assigned, and assume a generating process of the form
\begin{align}
\nn
\alpha_1, \alpha_2 &\sim p(\alpha)
\\
\nn 
N &\sim p(N)
\\
c_1 \ldots c_N &\sim p(c_{1},\ldots, c_{N}|\alpha_1) 
\label{eq:gen1}
\\
\mu_1 \ldots \mu_K|c_{1:N} &\sim p(\mu_1, \ldots \mu_K |\alpha_2) 
\nn
\\
x_i &\sim p(x_i|\mu_{c_{i}}) \quad i=1 \ldots N.
\nn
\end{align}
Here $\alpha_1, \alpha_2$ are hyperparameters. 
The number of clusters~$K$ is a random variable, indicating the number of distinct values among the sampled $c_i$'s, and~$\mu_k$ denotes a parameter vector controlling the distribution of the $k$-th cluster (e.g., $\mu_k$ could include both the mean and covariance of a Gaussian 
mixture 
component).
We assume that the priors $p(c_{1:N}|\alpha_1)$ and 
$p(\mu_{1:K}|\alpha_2)$ are exchangeable,
\eqan 
\nn
p(c_{1}, \ldots, c_N|\alpha_1) = 
p(c_{\sigma_1},\ldots, c_{\sigma_N}|\alpha_1)\,,
\enan
where $\{ \sigma_i \}$ is an arbitrary permutation of the indices,
and similarly for $p(\mu_{1:K}|\alpha_2)$.
Our interest in this work is in cases where $K$ 
can take any value $K \leq N$, such as the Chinese Restaurant Process~(CRP)
or its Pitman-Yor 
generalization~
{(see~\citet{npbreview} for a review).} 
Of course, our methods will also work for 
models with $K<B$ with fixed $B$, such as Mixtures of Finite Mixtures~\citep{miller2018mixture}.

Instead of representing configurations using $N$ labels~$c_i$, an alternative 
is obtained using 
$K$ sets of {\it indices}:
\eqan 
\s_k &=& (s_{k,1}, \ldots, s_{k,N_k}) \qquad k=1\ldots K \,,
\label{sk}
\\
\nn
 \textrm{where}  && \forall k, \forall i, c_{s_{k,i}} = k.
\enan 
{For example, the labels $c_{1:6}=(1,1,2,1,2,1)$
are equivalent to $\s_1=(1,2,4,6)$, 
$\s_2=(3,5)$.}
Note that cluster $k$ has size $N_k$ and $N = \sum_{k=1}^K N_k$. 
Given $N$ data points $\mathbf{ x} = \{x_i\}$, 
we would like to draw independent samples from the posterior $p(\cc|\x)$. For this, 
we consider expanding $p(\cc|\x)$ using either the labels
representation,
\eqan 
p(c_{1:N}|\mathbf{x}) 
=  p(c_1| \mathbf{ x} ) p(c_2|c_1,\mathbf{ x} ) \ldots p(c_N|c_{1:N-1}, \mathbf{ x}),
\label{joint}
\enan
or the indices representation,
\eqan 
p(\s_{1:K}|\x) = 
p(\s_1|\x) p(\s_2|\s_1,\x) \ldots p(\s_K|\s_{1:K-1}, \x) \,.
\label{psk}
\enan


Note that for a given cluster configuration,
$p(c_{1:N}|\mathbf{x}) = p(\s_{1:K}|\x)$. 
In the next two Sections, 
we present neural architectures 
to model the factors 
in each of these expansions. 


\section{Pointwise Sampling}
\label{sec:NCP}
We would like to model all the factors in (\ref{joint}) in a unified way, with a generic factor given~by 
\eqan
p(c_n|c_{1:n-1}, \mathbf{ x}) 
= \frac{p(c_1 \ldots c_n, \mathbf{ x})}
{ \displaystyle \sum_{c_n'=1}^{K+1}  p(c_1 \ldots c_n', \mathbf{ x})}.
\label{conditional}
\enan 
Here we assumed that there are $K$ unique values in $c_{1:n-1}$,
and therefore $c_n$ can take $K+1$ values, corresponding to $x_n$ joining any of the $K$ existing clusters, or forming its own new 
cluster.

Since (\ref{conditional})
is in general difficult to compute directly,
we will approximate these terms with a neural network 
$q_{\theta}(c_n| c_{1:n-1}, \mathbf{ x})$,
that takes as inputs $(c_{1:n-1}, \mathbf{ x})$, then extracts features and  combines them nonlinearly to output a probability distribution on $c_n$. 
Critically, we will design the network to enforce 
the highly symmetric structure
of (\ref{conditional}). 

To make this symmetric structure more transparent,
let us consider the 
joint distribution of the assignments of the first~$n$ data points,
\eqan 
p(c_1, \ldots, c_n, \mathbf{ x}) \,.
\label{joint_n}
\enan 
Note that under the 
model~(\ref{eq:gen1}),
this quantity depends 
on all the $N$ elements of $\x$, 
not just on $x_{1:n}$.
A neural representation of (\ref{joint_n}) should respect the permutation symmetries 
imposed on the~$x_i$'s by the values of~$c_{1:n}$. Therefore, our first task is  to build
permutation-invariant representations of the observations $\x$.  
The general problem of constructing such invariant encodings was discussed recently in \citep{deep_sets}; to adapt this approach to our context, we consider three distinct permutation symmetries:

\newlist{myitemize}{itemize}{3}
\setlist[myitemize,1]{label=\textbullet,leftmargin=0.2in}

\begin{myitemize}
	
	\item {\bf Permutations within a cluster: }	
	(\ref{joint_n}) is invariant under permutations of $x_i$'s in the same cluster.
	For each of the $K$ clusters that have been sampled so far, we define the encoding 
	\eqan
	H_k= \sum_{i : c_{i}=k} h(x_i)  
	\quad h:\mathbb{R}^{d_x} \rightarrow \mathbb{R}^{d_h}
	\label{Hk}
	\enan 
	which is clearly invariant under permutations of $x_i$'s in the same cluster. In general $h$ is an encoding function we learn from data, unless	$p(x|\mu)$	belongs to an exponential family and the prior $p(c_{1:N})$ is constant, as discussed in SM Section B. 


	\item {\bf Permutations between clusters: }
	(\ref{joint_n}) is invariant under permutations of the cluster labels. 		
	In terms of the within-cluster invariants~$H_k$, this can be captured by 	
	\eqan 
	G= \sum_{k =1}^K g(H_k) ,
	\quad 
	g:\mathbb{R}^{d_h} \rightarrow \mathbb{R}^{d_g}.
	\label{G_def}
	\enan 			

	\item {\bf Permutations of the unassigned data points: }
	(\ref{joint_n}) is also invariant under permutations of the $N-n$ 
	unassigned data points. 
	This can be captured by 
	\eqan 
	U = \sum_{i=n+1}^{N}  u(x_i) \,,
		\qquad 
	u:\mathbb{R}^{d_x} \rightarrow \mathbb{R}^{d_u}.
	\label{Q}
	\enan 		
	
\end{myitemize}

\begin{figure}[t!]
	\begin{center}
		\fbox{		
		\includegraphics[width=.45\textwidth]{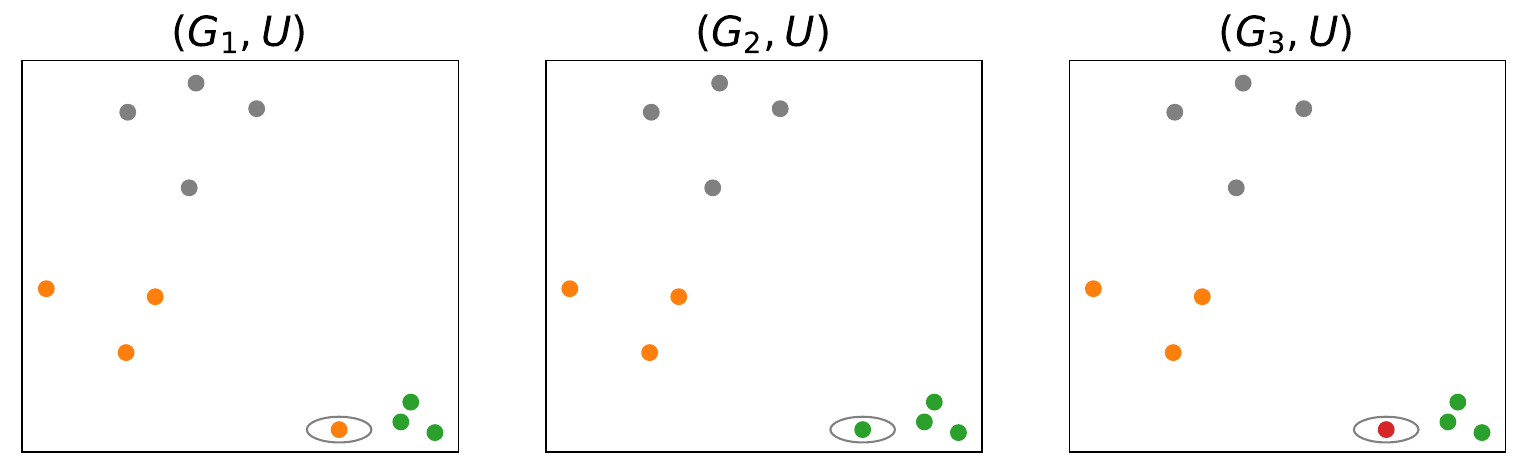}}
	\end{center}
	\caption{{\bf Encoding cluster labels.} 
After assigning labels 
$c_{1:6}$ to $K=2$ clusters, each of the three possible $c_7$
labels  (for the circled point~$x_7$) gives
an encoding  $G_k$ for the set $x_{1:7}$.  
The vector~$U$ encodes the four gray unlabeled points (Best in color).
	} 
	\label{fig:Gks}	
\end{figure}

Note that $G$ and $U$ provide fixed-dimensional, 
symmetry-invariant representations of the assigned and non-assigned data points, respectively,
for any values of  $N$ and $K$.  
Encodings of this form 
yield arbitrarily accurate approximations 
of (partially) symmetric 
functions~\citep{deep_sets,gui2019pine}.


\subsection{The Variable-input Softmax}
After assigning values to $c_{1:n-1}$,
each of the $K+1$ possible values for $c_n$ 
corresponds to $h(x_n)$ appearing in 
one particular  $H_k$ in (\ref{Hk}), and 
yields a separate vector $G_k$ in (\ref{G_def}). 
See Figure~\ref{fig:Gks} for an example.
In terms of the $G_k$'s and $U$, we propose to model~(\ref{conditional}) as
\eqan 
q_{\theta}(c_n=k|c_{1:n-1}, \mathbf{ x}) = \frac{ e^{f(G_k,U)} }
{  \sum_{k'=1}^{K+1} e^{f(G_{k'},U)}       } 
\label{ncp}
\enan
with $k = 1\ldots K+1$, where we have introduced a new real-valued function $f$. 
In other words, each value of $c_n$ corresponds
to a different channel through which the encoding $h(x_n)$ flows to the logit value $f$. 
Note that $k=K+1$ corresponds to $c_n$ forming its own new cluster with $H_k=h(x_n)$.

Our softmax~(\ref{ncp}) differs from the
usual form in, e.g., classification networks, 
where a fixed number of categories receive  
logit values $f$ from the fixed-size final layer of a multi-layer perceptron (MLP). In our case, the 
discrete identity of each logit is determined by the neural path that the 
input $h(x_n)$ takes to $G$, thus allowing a flexible number of categories.


In eq.~(\ref{ncp}),
$\theta$ denotes the parameters in the functions $h,g,u$ and $f$,
which we represent with neural networks. 
By storing and updating $G$ and $U$ for successive values of~$n$, 
as shown in Algorithm 1, 
the computational 
cost of a full i.i.d.\ sample of $c_{1:N}$ is $O(NK)$, 
the same as a single Gibbs sweep; 
and by parallelizing steps 8-9
in Algorithm 1, the number of 
network forward passes 
becomes $O(N)$.
We term this approach Neural Clustering Process (NCP).
It is relatively easy to run
hundreds of copies of Algorithm~1 in parallel
on a GPU, with each copy yielding a different set of 
samples~$c_{1:N}$.\footnote{Implementation available at 
\url{https://github.com/aripakman/neural_clustering_process}}


  \begin{algorithm}[t!]
	\caption{$O(NK)$  Neural Clustering Process}
	\label{algo}
	\begin{algorithmic}[1]	
		\STATE $h_i \gets h(x_i), \, u_i \gets u(x_i)   \qquad i=1 \dots N$  \hfill \COMMENT{Notation}
		\STATE $U \gets \sum_{i=2}^N u_i$, \, $K \gets 1$  \hfill \COMMENT{Initialize unassigned set}
		\STATE $H_1 \gets h_1$, $G \gets g(H_1)$, $c_1 \gets 1$ 
		\hfill \COMMENT{First cluster}
		\FOR{$n\gets 2\ldots N$}
		\STATE $U \gets U - u_n$  \hfill \COMMENT{Remove $x_n$ from unassigned set}
		\STATE $H_{K+1} \gets 0$   \hfill \COMMENT{We define $g(0)=0$}
		\FOR{$k\gets 1\ldots K+1$}  
		\STATE $G_k \gets G +g(H_k + h_n) - g(H_k)$ \hfill \COMMENT{Add $x_n$}
		\STATE $q_{k} \gets e^{f(G_k,U)}$ 
		\ENDFOR
		\STATE $q_k \gets q_k/\sum_{k'=1}^{K+1}q_{k'}$, \,\, $c_{n} \sim q_k$
		\hfill \COMMENT{Sample}
		\IF{$c_{n} = K+1$}
		\STATE $K \gets K+1$
		\ENDIF 
		\STATE $G \gets G-g(H_{c_{n}}) + g(H_{c_{n}} + h_{n})$  \hfill \COMMENT{Add point $x_{n}$}
		\STATE $H_{c_{n}} \gets H_{c_{n}} + h_{n}$
		\ENDFOR
		\STATE Return $c_1 \ldots c_N$			
	\end{algorithmic}
\end{algorithm}

\subsection{Objective Function}
In order to train 
the neural networks, 
we use stochastic 
gradient descent to minimize the 
expected KL divergence,
\eqan 
& \mathbb{E}_{p(N)p(\mathbf{x})} 
KL ( p(c|\mathbf{ x}) \Vert
q_{\theta}(c|\mathbf{ x}) )
= 
\label{eq:KL_NCP}
\\
\nn
& -    
\mathbb{E}_{p(N) p(c_{1:N},\mathbf{ x})}
\left[  \sum_{n=2}^{N} \log q_{\theta}(c_{n} |c_{1:{n-1}},\mathbf{x})
\right] + \textrm{const.}
\enan 
Samples from $p(c_{1:N},\mathbf{ x})$ are obtained from the generative model,
irrespective of the model being conjugate. 
In cases with unlimited samples (such as the 2D Gaussian example in~\cref{sec:examples} and the spike-sorting application in~\cref{sec:spike_sorting}), we can potentially train a neural network to approximate $p(c_n|c_{1:n-1},\x)$ arbitrarily accurately.

The objective function (\ref{eq:KL_NCP}) 
can be seen as a form of Expectation Propagation~\cite{minka2001expectation}, as opposed to variational inference, 
which would minimize instead 
$KL( q_{\theta}(c|\mathbf{ x}) 
\Vert p(c|\mathbf{ x})  )$. Note that the gradient acts only on the variable-input 
softmax~$q_{\theta}$, not on
$p(c,\x)$, so there is no problem 
of backpropagating through discrete variables \citep{jang2016categorical, maddison2016concrete}.

\section{Clusterwise Sampling}
\label{sec:CCP}
While the NCP algorithm 
is good enough for 
small datasets,
$O(N)$ forward calls 
might be too many for large datasets. We consider  now an $O(K)$ alternative, based on modeling the factors 
in the clusterwise expansion~(\ref{psk}),
\eqan 
p(\s_{1:K}|\x) = 
p(\s_1|\x) p(\s_2|\s_1,\x) \ldots p(\s_K|\s_{1:K-1}, \x) \,.
\label{eq:psk2}
\enan 


Sampling from  $p(\s_k|\s_{1:k-1}, \x )$ can be done in two steps: 
\begin{enumerate}
    \item Sample uniformly an index $d_k$  from the set $I_k = \{1\ldots N\} \backslash  \{\s_{1:k-1}\}$ of 
available indices (those not taken by $\s_{1:k-1}$). The point $x_{d_k}$ 
becomes the first element of cluster $k$.
\item
Denote by $\ba_k = (a_1 \ldots a_{m_k})$ 
the elements of the set of remaining indices
$I_k \backslash \{d_k\}$,
where  $m_k = |I_k \backslash \{d_k\}|$.
Conditioned on $(d_k, \s_{1:k-1}, \x)$, 
sample a binary vector 
\eqan 
\bb_{k} = (b_1 \ldots b_{m_k}) \qquad \in \{0,1\}^{m_k}
\nn
\enan 
with $b_i=1$ if the point $x_{a_i}$ 
joins cluster $k$. 
\end{enumerate}

\begin{figure}[t!]
	\begin{center}
		\fbox{		
		\includegraphics[width=.45\textwidth]{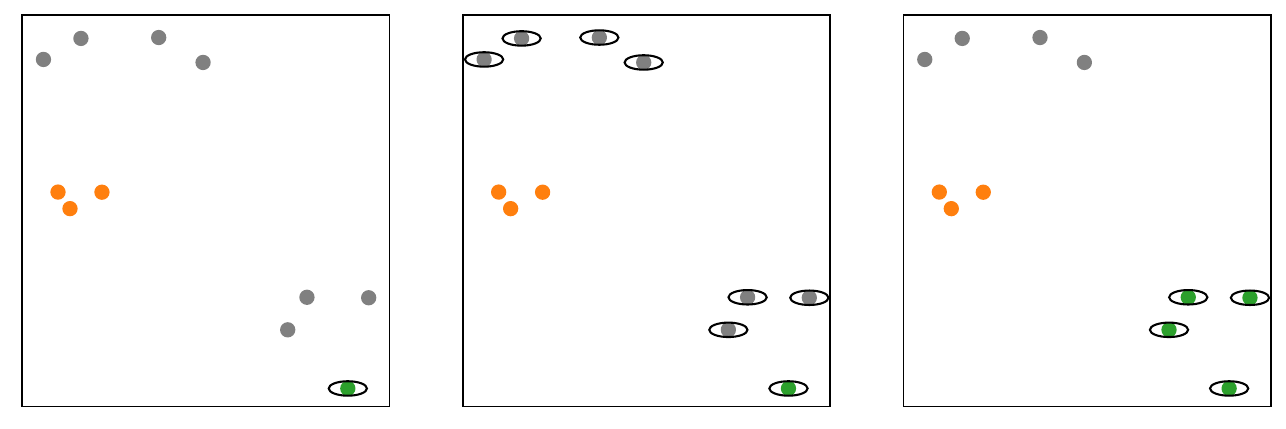}}
	\end{center}
	\caption{{\bf Clusterwise sampling.}  
	{
	{\it Left}: After sampling cluster $\s_1$ (orange),
	the first element of $\s_2$, $d_2$, is sampled uniformly (green). 
	{\it Middle:}~All unassigned points $\ba_2$ (grey)
	are  candidates 
	to join~$d_2$.
	{\it Right:} By sampling 
	$\bb_{2}$, cluster $\s_2$ is completed. 
 (Best in color).}
	} 
	\label{fig:CCP}	
\end{figure}

These two 
steps (see~\cref{fig:CCP} for an example) are iterated until there are no available indices left, and
have probability
\eqan 
p(d_k,\bb_{k} | \s_{1:k-1},\x) = p(d_k|\s_{1:k-1})p(\bb_{k}|d_k,\s_{1:k-1},\x) 
\label{eq:two_factor}
\enan 
where
\eqan 
\nn
p(d_k|\s_{1:k-1}) = \left\{
\begin{tabular}{cc}
    $1/ |I_k|$ & \textrm{for} $d_k \in I_k$ ,
    \\
    0 &  \textrm{for} $d_k \notin I_k$  ,
\end{tabular}
\right.
\enan 
and $|I_k|=m_k+1$.
The event indicated by $\s_k$
is  the union of  $N_k$ 
disjoint  events $(d_k,\bb_k)$, 
and we have 
\eqan
p(\s_k|\s_{1:k-1},\x) 
= \frac{1}{|I_k|} {\scriptstyle 
\sum_{d_k \in \s_k}}
p(\bb_{k}|d_k,\s_{1:k-1},\x)
\label{mix_model}
\enan 
where $\bb_{k}$ has a `1' for each 
element in $\s_k$ except $d_k$.
Our major challenge is 
therefore to model the conditional $p(\bb_{k}|d_k,\s_{1:k-1},\x)$,
which we address next. 
 
 \begin{figure*}[t!]
	\begin{center}
		\includegraphics[width=\textwidth, height=1.1in]{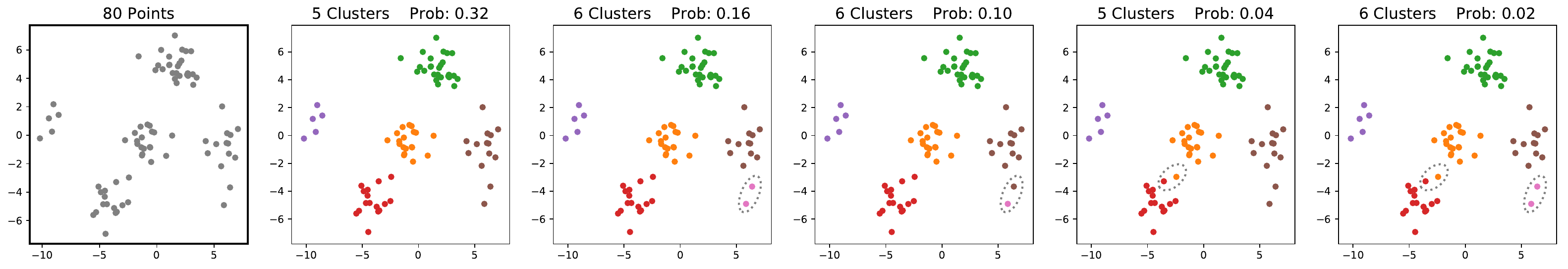}	
	\end{center}
	\vspace{-2mm}
	\caption{
		{
			{\bf Mixture of 2D Gaussians:} 
			Given the observations in the first panel, 
			we show samples from the NCP posterior. 
			Note that less-reasonable samples are assigned lower probability by the NCP. The dotted ellipses indicate departures from the first, highest-probability sample.
			Our GPU implementation gives thousands of samples in less than a second. 
			CCP results are similar. 
			(Best in color.)
		}
	}
	\label{fig:exact_vs_ncp}
	
\end{figure*}

\subsection{Factorized posterior}

The information contained in $(d_k,\s_{1:k-1},\x)$,
is better represented
by splitting the dataset as 
$\x_k = (\x_{\bfa},x_{d_k},\x_{\s})$,
where
\begin{center}
\begin{tabular}{ll}
    $\x_{\bfa}$ = $ (x_{a_1} \ldots x_{a_{m_k}})$ 
    & $m_k$ available points for  cluster k 
    \\
    $x_{d_k}$&  First data point  in cluster k 
    \\
    $\x_{\s}=(\x_{\s_1} \ldots \x_{\s_{k-1}} )$&   
    Points already assigned to clusters. 
\end{tabular}
\end{center}
Thus $p(\bb_k|\x_k)\equiv p(\bb_{k}|d_k,\s_{1:k-1},\x)$. 
Note now that this factor
has a form of conditional exchangeability
\eqan
\nn 
&& p(b_{1} \ldots b_{m_k} | 
x_{a_1},\ldots, x_{{a_{m_k}}}, x_{d_k},\x_{\s}  )
= 
\\
\nn
&& \qquad 
p(b_{\sigma_{1}} \ldots b_{\sigma_{m_k}} |x_{\sigma_{a_1}} \ldots x_{\sigma_{a_{m_k}}} , x_{d_k},\x_{\s}  )\,,
\enan 
where $\sigma$ is an arbitrary permutation of the 
elements of $\bb_k$ and $\x_{\bfa}$. 
Based on this, we assume a conditional version of de Finetti's theorem and propose\footnote{More precisely, de Finetti's theorem~\cite{bruno1931,hewitt1955symmetric} holds for infinite sequences.
For finite sequences, as in our case, 
the result has been shown to 
hold 
only approximately
and for discrete variables, 
both in the unconditional~\cite{diaconis1977finite,diaconis1980finite} and conditional cases~\cite{christandl2009finite}.
} 
\eqan
p(\bb_k| \x_k) \simeq \int \!\!
d\z_k  \!
\prod_{i=1}^{m_k} 
p_i(b_{i}|\z_k,\x_k)  
p(\z_k|\x_k) \,,
\label{eq:marginal_z}
\enan 
and approximate  the integrands as 
\eqan 
p_{\theta}(\z_k|\x_k) &=& 
 {\cal N}(\z_k|\x_k)
\label{eq:pz_x}
\\
p_{\theta,i}(b_{i}|\z_k, \x_k)
&=& \textrm{sigmoid}[\rho_i(\z_k, \x_k)] \,.
\label{eq:pb_zx}
\enan 
Crucially, the posterior distributions of the  $b_{i}$'s are conditionally independent. Therefore, after sampling $p(\z_k|\x_k)$, all the $b_{i}$'s can be sampled in parallel.
Thus, while a full sample of~(\ref{eq:psk2}) 
of course has cost $O(N)$, the heaviest
computational burden, from network evaluations, 
scales as $O(K)$, since each
factor in~(\ref{eq:psk2}) needs $O(1)$ forward calls. As in NCP,
we can get hundreds of 
full samples via GPU parallelization.

To summarize, the elements of $\s_k$ are generated in a process with latent variables $d_k,\z_k$ 
and joint distribution 
\eqan 
\quad & p_{\theta}(\s_k,\z_k,d_k|\s_{1:k-1},\x)=
\nn
\\
\nn 
&p_{\theta}(\bb_k|\z_k, \x_k) p_{\theta}(\z_k|\x_k)p(d_k|\s_{1:k-1})
\\
\text{where} & 
\nn
\\
& p_{\theta}(\bb_k|\z_k, \x_k)= \prod_{i=1}^{m_k}
p_{\theta,i}(b_{i}|\z_k, \x_k)
\,.
\label{eq:pb}
\enan 
In order to learn these functions, we
introduce 
an encoder 
$q_{\phi}(\z_k,d_k|\s_{1:k}, \x)$
to approximate the intractable 
posterior, 
and train the functions as a 
conditional variational autoencoder (VAE)
~\cite{sohn2015learning} 
(as we condition everything on  $\x$).
The dependence of all the functions 
on the components of~$\x$ should respect the symmetries imposed 
by the conditioning  clusters $\s_{1:k-1}$ (or $\s_{1:k}$
for $q_{\phi}$).
This can be achieved 
using encodings 
similar to those we used above 
in~\cref{sec:NCP};
see SM Section~A for details.

Let us stress the double role 
of 
$p_{\theta}(\z_k|\x_k)
p(d_k|\s_{1:k-1})$ and $p_{\theta}(\bb_k|\z_k,\x_k)$. In the VAE framework, they are
the priors and likelihood of a 
generative model for~$\s_k$.
On the other hand they represent, after $d_k,\z_k$ marginalization~(\ref{mix_model})-(\ref{eq:marginal_z}), a factor of the posterior expansion~(\ref{eq:psk2}). 
We call this approach Clusterwise Clustering 
Process~(CCP).

\begin{figure*}[t!]
	\centering
	\fbox{\includegraphics[height=5cm, width=\textwidth]{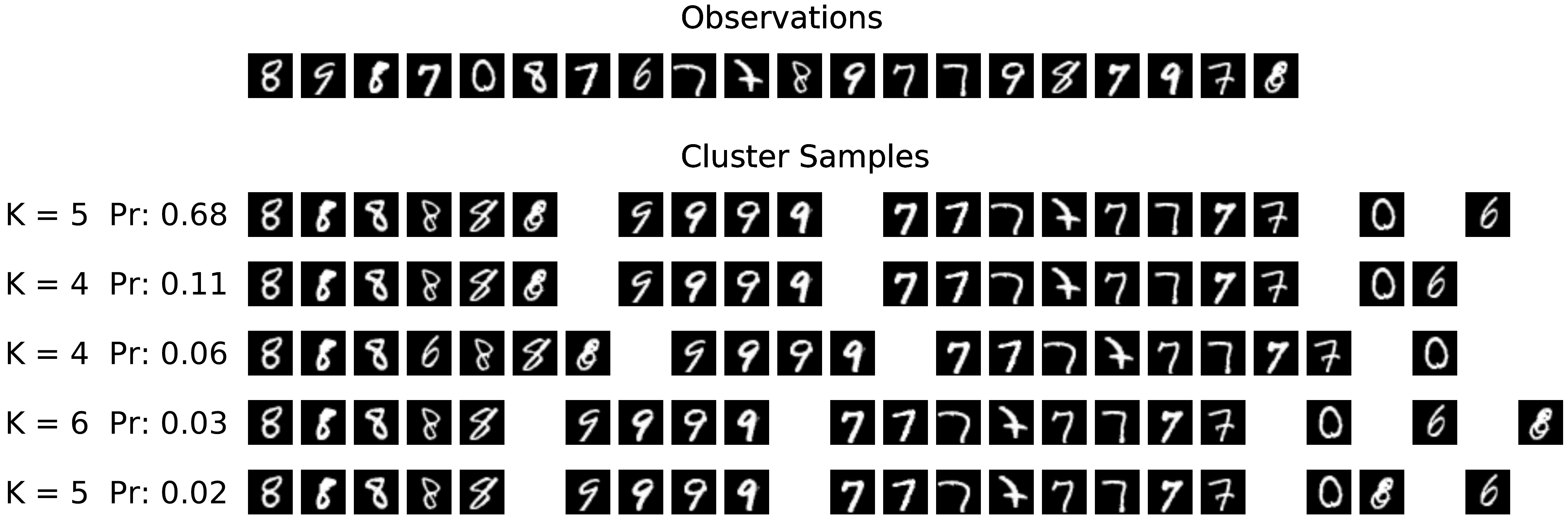}
	}
	\vspace{-2mm}
	\caption{{\bf NCP trained on MNIST clusters.} 
	Top row: 20 images from the MNIST test set. 
	Below: five samples of $c_{1:20}$ from the NCP posterior.
    Note that each sample captures 
some ambiguity suggested by the form of particular digits. 
CCP results are similar. 
	}
	\label{fig:mnist}
\end{figure*}

\subsection{Objective Function}
Similar to  the  NCP case in (\ref{eq:KL_NCP}), we 
want an approximation $p_{\theta}(\s_{1:K}|\x)$ to $p(\s_{1:K}|\x)$ that
maximizes 
\eqan 
& -\mathbb{E}_{p(\x)} KL[ p(\s_{1:K}|\x) | |p_{\theta}(\s_{1:K}|\x) ] 
\label{eq:ccp_objective}
\\
& = \mathbb{E}_{p(\x, \s_{1:K})}
\sum_{k=1}^K
\log p_{\theta}(\s_{k}|\s_{1:k-1}, \x)
+ \textrm{const.}
\nn
\enan 
where  
we expanded $p_{\theta}(\s_{1:K}|\x)$
as in (\ref{eq:psk2}).
Using now the variational posterior $q_{\phi}$, we can bound (\ref{eq:ccp_objective}) from below, which 
leads us to maximize the ELBO 
\eqan
\nn
{
\mathbb{E}_{p(\x,\s_{1:K})} 
\sum_{k=1}^K 
\mathbb{E}_{q_{\phi}(\z_k,d_k|\s_{1:k},\x)}
\log \left[
\frac{p_{\theta}(\s_k,\z_k,d_k|\s_{1:k-1},\x)}
{q_{\phi}(\z_k,d_k|\s_{1:k},\x)}
\right]}
\enan 
To use the reparametrization trick~\cite{kingma2013auto}, we use a Gumbel-Softmax  relaxation 
for $d_k$~\citep{jang2016categorical,maddison2016concrete}.
See SM Section A.

\subsection{Estimating sample probabilities}
Unlike NCP, CCP samples do not come with a probability estimate. 
The latter can be estimated using (\ref{eq:psk2}) and 
\eqan 
p(\bb_k|\x_k) &\simeq& 
\frac{1}{M} 
\sum_{j=1}^M 
p_{\theta}(\bb_k|\z_{k,j}, \x_k)
\enan 
where $\z_{k,j} \sim p_{\theta}(\z_k|\x_k )$.

\section{Examples}
\label{sec:examples}

{\bf 2D Gaussian models:}
The generative model is 
\eqan 
\begin{aligned}[c]
\alpha &\sim \textrm{Exp}(1) \quad c_{1:N}  \sim \textrm{CRP}(\alpha) 
\\
N &\sim \textrm{Uniform} [5,100]
\end{aligned}
\quad
\begin{aligned}[c]
\mu_k &\sim N(0,\sigma_{\mu}^2  \mathbf{1}_2)  
\label{2D_gauss}
\\
x_i &\sim N(\mu_{c_{i}}, \sigma^2  \mathbf{1}_2) 
\end{aligned}
\nn
\enan
where CRP stands for the Chinese Restaurant Process, 
with concentration parameter $\alpha$, $\sigma_{\mu}=10$, and $\sigma=1$.
\cref{fig:exact_vs_ncp} shows that the NCP  captures the posterior uncertainty inherent in clustering this data. 
Since we have unlimited samples,
there is no distinction here between training and test sets.

{\bf MNIST digits:} 
We consider next a DPMM over MNIST digits, with generative model
\begin{equation*}
\begin{aligned}[c]
\alpha &\sim \textrm{Exp}(1) \quad c_{1:N}  \sim \textrm{CRP}_{10}(\alpha) 
\\
N &\sim \textrm{Uniform} [5,100]
\\
l_k &\sim \textrm{Unif} [0,9]  - \textrm{without replacement.}   \quad k=1 \ldots K
\nn
\\
x_i &\sim \textrm{Unif} [\textrm{MNIST digits with label } l_{c_i}] 
\quad i=1 \ldots N
\end{aligned}
\end{equation*}
where $\textrm{CRP}_{10}$
is a Chinese Restaurant Process 
truncated to up to 10 clusters,
and $d_x=28\times28$. 
Training was performed by 
sampling $x_i$ from the MNIST training set. 
\cref{fig:mnist} shows posterior samples 
for a set of digits from the MNIST test set,
illustrating how the estimated model correctly captures the shape ambiguity of some of the digits.
Note that in this case the generative model has no analytical expression, 
but this presents no problem; a set of labelled samples is all we need for training.
See SM Section G for details of  the 
network architectures used.

\section{Related works}
\label{sec:related}
Most works on neural network-based clustering
focus on learning features 
as inputs to traditional clustering algorithms,
as reviewed in~\citep{du2010clustering,aljalbout2018clustering,min2018survey}. 
Our approach differs from these works because it leverages deep learning to improve 
{\it algorithmic} aspects of clustering, via amortization.

Permutation-invariant neural architectures have been explored recently 
in~\citep{pmlr-v70-ravanbakhsh17a,BRUNO,
lee2018set,BloemReddy,wagstaff2019limitations}.
The representation of a set via a sum (or mean) of encoding vectors was also used in~\citep{guttenberg2016permutation,ravanbakhsh2016deep,	edwards2016towards,
deep_sets,garnelo2018conditional,kim2019attentive}.

A conditional form of de Finetti's theorem
was also assumed for Neural Processes (NP)~\cite{garnelo2018neural},
but differs from our assumed form  in (\ref{eq:marginal_z}) 
in that our prior $p_{\theta}(\z_k|\x_k)$ depends symmetrically
on the available points~$\x_{\ba}$, 
in order to keep the correct dependency 
of the marginal $p(\cc_{1:n},\x)$
on all the $N$ components of $\x$, 
while for NPs the prior 
is independent of the 
available data points.

Amortized inference of Gaussian
mixtures
has been studied recently 
in~\citep{le2016inference,lee2018set,kalra2019learning}.
In these works the output of the network are the 
mixture parameters instead of 
 sampled discrete labels,
and the number of components is either 
bounded~\citep{le2016inference} or fixed~\citep{lee2018set,kalra2019learning}.
Closer to  our CCP  is the DAC approach~\cite{DAC},
that uses the set attention mechanism 
of~\cite{lee2018set} in the encoder to iteratively  isolate and eliminate one cluster per iteration, 
in  $O(K)$ network evaluations. 
But the  clusters have no clear interpretation in terms of the generative model, as they come
from hard thresholding of 
sigmoids and the eliminated clusters do not appear as 
a conditioning context to find new clusters.
 We summarize these comparisons in~\cref{table:compare}.



\begin{table}[h!]
\begin{center}
\begin{tabular}{|l|c|c|c|c|}
\hline 
Property  & {\scriptsize CCP} & { \scriptsize NCP} &  
{ \scriptsize DAC} & { \scriptsize MoG} 
\\ 
\hline 
Unlimited components   
& 
$\checkmark$  & $\checkmark$ & $\checkmark$ & \xmark 
\\
Amortized labels  &
$\checkmark$ & $\checkmark$ & $\checkmark$ & \xmark 
\\
Any generative model &
$\checkmark$ & $\checkmark$ & $\checkmark$ & \xmark 
\\
Well defined posterior &
$\checkmark$ & $\checkmark$ & \xmark & - 
\\
Forward passes  &
$\scriptstyle O(K)$ & $\scriptstyle O(N)$ & $\scriptstyle O(K)$ & $\scriptstyle O(1)$
\\
\hline 
\end{tabular}
\caption{ {\bf Comparing amortized clustering approaches.}
We compare NCP/CCP (our methods) 
with DAC~\citep{DAC} and amortization for mixtures of Gaussians (MoG) ~\citep{le2016inference,lee2018set,kalra2019learning}.
}
\label{table:compare}
\end{center}
\end{table}

\begin{figure*}[t!]
\includegraphics[width=.96\textwidth,height=1.3in]{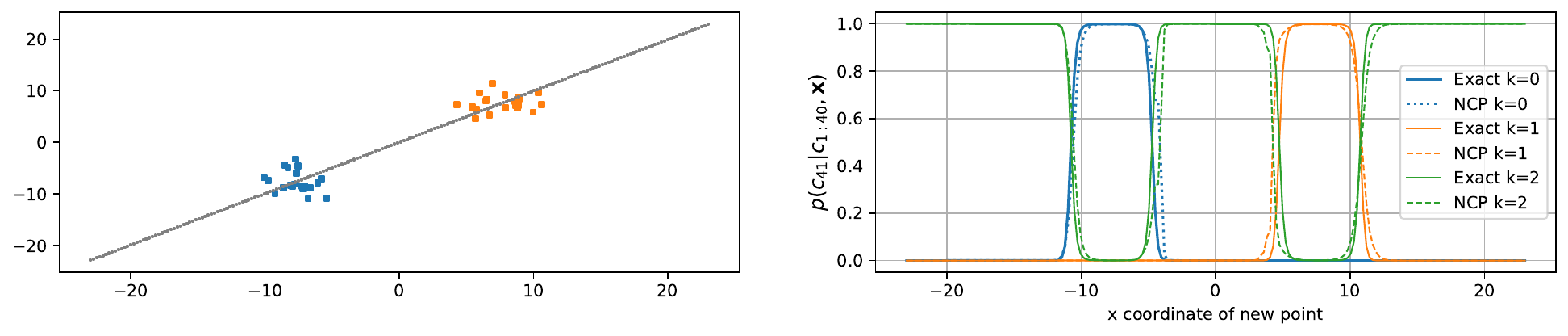}
\includegraphics[width=.96\textwidth,height=1.4in]{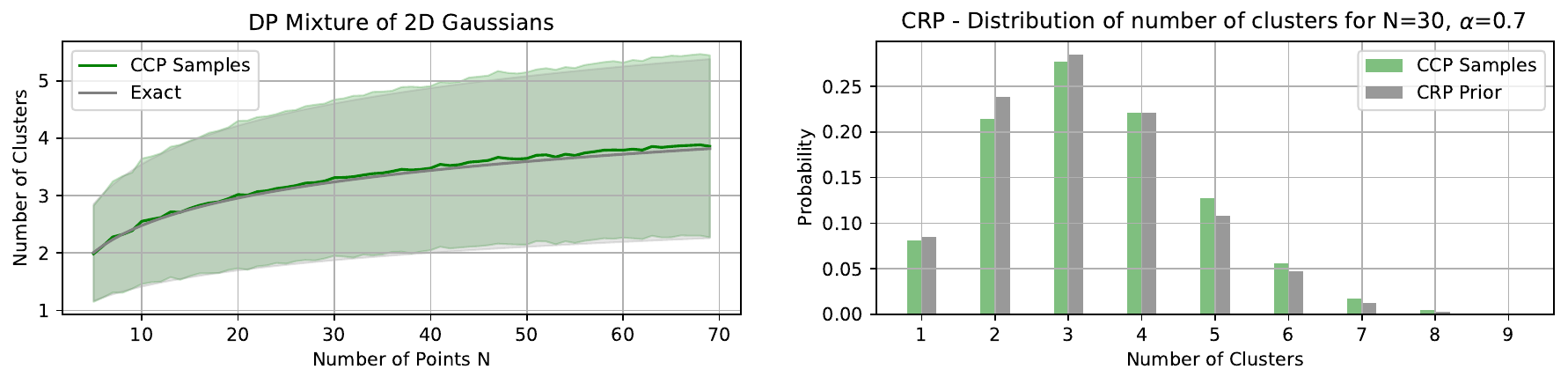}	

	\vspace{-2mm}
	\caption{{\bf Quantitative Evaluations.} 
				{\it Upper left:} Two clusters of 20 points each  and a line over possible locations of a 41st last point. 
			{\it Upper right:} Assuming the 2D model from (\ref{2D_gauss}), the posterior $p(c_{41}|c_{1:40}, \mathbf{ x})$ 
			can be computed exactly, and we compare it to the NCP estimate as a function of 
			the horizontal coordinate of $x_{41}$, as this point moves over the gray line on the upper left panel.
		{\it Geweke's Tests. Lower left: }
		The curves compare the exact mean ($\pm$~std.) of the number of clusters $K$ for different~$N$'s from the CRP prior ($\alpha=0.7$),
		with CCP sampled estimates using eq.~(\ref{geweke}).
		{\it Lower right: }
		Similar comparison for the 
		 histogram of $K$ for $N=30$  points.
	}	
	\label{fig:geweke}
\end{figure*}

\section{Evaluations and diagnostics}
\label{sec:QE}


The examples in~\cref{sec:examples} provide strong qualitative evidence that our approximations to the true posteriors  in these models capture the uncertainty inherent in the observed data.  But we would like to go further and ask quantitatively how well our approximations match the exact posterior.  Unfortunately, for sample sizes much larger than $N = O(10)$ it is impossible to compute the exact posterior in these models.  Nonetheless, there are several quantitative metrics we can examine to check the accuracy of the model output. 
{Note that the diagnostics below that rely on the probabilistic 
nature of the inferred clusters
are not applicable to the other methods
compared in Table~\ref{table:compare}.}


{\bf NCP vs. CCP: }
The results from the two approaches 
were similar in all the examples we considered, 
such as those 
in~\cref{sec:examples}. Training CPP, however, presents the usual challenges 
of VAEs. We found it useful to use 
multiple sample objectives~\cite{burda2015importance} and estimate the gradient 
using double-reparametrization~\cite{tucker2018doubly}.

{\bf Global symmetry from exchangeability:}
From the exchangeability of  
$p(c_{1:N}|\alpha_1)$, the expansion (\ref{joint}) 
should not depend on the order of the data points,
but this symmetry is not enforced explicitly.
If our model learns the conditional probabilities correctly, this symmetry should be (approximately) satisfied, 
as we show in SM Section~C.

{\bf Estimated vs. Analytical Probabilities:}
Some conditional probabilities can be computed analytically and compared  with the estimates output by the network; in the example shown in Figure~\ref{fig:geweke}, upper-right, the estimated probabilities are in close agreement with their exact values.

{\bf Geweke's Tests:}
A popular family of tests 
that check the 
correctness of MCMC implementations~\citep{geweke2004getting}  
can also be applied in our case: verify the (approximate) identity between 
the prior $p(c_{1:N})$ and 
\begin{equation}
q_{\theta}(c_{1:N}) \equiv \int d\x \, q_{\theta}(c_{1:N}|\x) \, p(\x) \,,
\label{geweke}
\end{equation}
where $p(\x)$ is the marginal from the generative model. 
Figure~\ref{fig:geweke} shows such a comparison for the
2D Gaussian DPMM from~\cref{sec:examples}, showing excellent agreement. 


{\bf Comparison with~MCMC:}
NCP/CCP have some advantages over MCMC approaches.
First, unlike MCMC, 
we get a probability estimate for 
each sample, either directly (NCP) 
or with minimal computation (CCP). 
Secondly, NCP/CCP enjoy higher efficiency, due to 
parallelization of iid samples. For example, in the 
Gaussian 2D example in eq.(\ref{2D_gauss}), in the time a collapsed Gibbs sampler 
produces one (correlated) sample, 
 our GPU-based NCP implementation produces more than 100 iid approximate samples.
 Finally, NCP/CCP  do not need 
 a burn-in period.

{\bf Comparison with Variational Inference:}
Below  we compare NCP with a variational 
approach on spike sorting. For 2000 spikes, the 
latter returned one clustering estimate in 0.76 secs., but does not properly handle the uncertainty about the number of clusters. NCP produced 150 clustering configurations in 10 secs., efficiently capturing clustering uncertainty. In addition, the variational approach requires a preprocessing step that projects the samples to lower dimensions, whereas NCP directly consumes the high-dimensional data by learning an encoder function $h$.

\section{Application: spike sorting with NCP}
\label{sec:spike_sorting}

\begin{figure}[h!]
	\centering
	\includegraphics[width=.48\textwidth]{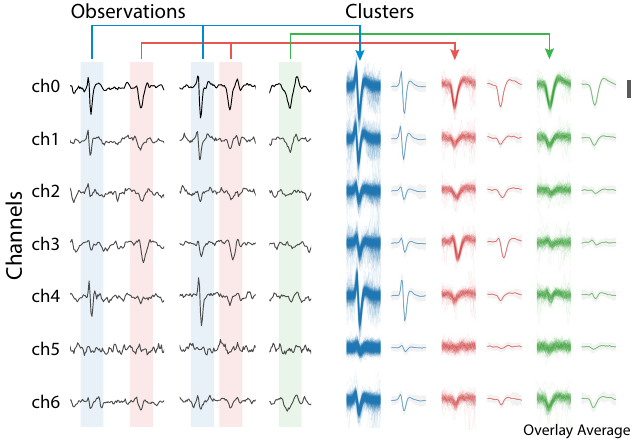}
	\vspace{-5mm}
	\caption{{\bf Clustering multi-channel spike waveforms using NCP.} Each row is an electrode channel. Spikes of the same color belong to the same cluster. (Scale bar: 5$\times$ noise s.d.).  
	}
	\label{fig:spike_sorting}
\end{figure}

Large-scale neural population recordings using multi-electrode arrays (MEA) are crucial for understanding neural circuit dynamics. 
Each MEA electrode reads the signals from many neurons, 
and each neuron is recorded by multiple nearby electrodes. 
As a key analysis step, spike sorting converts the raw signal into a set of spike trains belonging to individual neurons \citep{pachitariu2016kilosort, chung2017fully, Jun2016jrclust, lee2017yass, Chaure2018, carlson2019continuing}. 
At the core of many spike sorting pipelines is a clustering algorithm that groups the detected spikes into clusters, each representing a putative neuron (Figure \ref{fig:spike_sorting}). However, clustering spikes can be challenging: (1) Spike waveforms form highly
non-Gaussian clusters in spatial and temporal dimensions, and it is unclear what are the optimal features for clustering. (2) It is unknown \textit{a priori} how many clusters there are. (3) Existing methods do not perform well on spikes with low signal-to-noise ratios (SNR) due to increased clustering uncertainty, and fully-Bayesian approaches 
proposed to handle this uncertainty \citep{wood2008nonparametric,carlson2013multichannel} do not scale to large datasets. 

To address these challenges, we propose a novel approach to spike clustering using NCP. We consider the spike waveforms as generated from a Mixture of Finite Mixtures (MFM) distribution \citep{miller2018mixture}, which can be effectively modeled by NCP. 
(1) Rather than selecting arbitrary features for clustering, 
the spike waveforms are encoded with a convolutional neural network (ConvNet),
which is learned end-to-end jointly with the NCP network to ensure optimal feature encoding. 
(2) Using a variable-input softmax function, NCP is able to perform inference on cluster labels without assuming a fixed or maximum number of clusters. (3) NCP allows for efficient probablistic clustering by GPU-parallelized posterior sampling, which is particularly useful for handling the clustering uncertainty of ambiguous small spikes.
(4) The computational cost of NCP training can be highly amortized, 
since neuroscientists often sort spikes form many statistically similar datasets. 

We trained NCP for spike clustering using synthetic spikes from a simple yet effective generative model that mimics the distribution of real spikes, and evaluated the spike sorting performance on labeled synthetic data, unlabeled real data and hybrid test data by comparing NCP against two other methods: 
(1) {\bf  vGMFM},
    variational inference on Gaussian MFM \citep{hughes2013memoized}. 
%
(2) {\bf Kilosort}, a state-of-the-art spike sorting pipeline described in \cite{pachitariu2016kilosort}. 
In the Supplementary Material (SM) Section~D, we describe the dataset, neural architecture, and the training/inference pipeline of NCP spike sorting. 
{In SM Section~E, we show that NCP spike sorting achieves high clustering quality, and matches or outperforms a state-of-the-art method on synthetic, real and hybrid data.}

\begin{figure}[htb!]
    \centering
    \includegraphics[width=.48\textwidth]{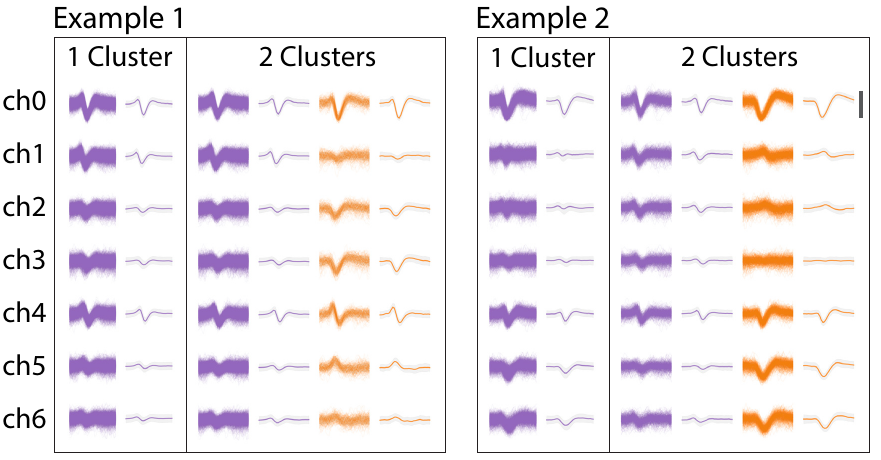}
    \vspace{-5mm}
    \caption{
      {\bf Clustering ambiguous small spikes.} In both examples, multiple plausible clustering results of small spikes were produced by sampling from the NCP posterior (scale bar = 5$\times$ noise s.d.).
      }
    \label{fig:spikesorting_smallspikes}
\end{figure}

\textbf{Probabilistic clustering of ambiguous small spikes.} Sorting small spikes has been challenging due to the low SNR and increased uncertainty of cluster assignment. 
By efficient GPU-parallelized posterior sampling of cluster labels, NCP is able to handle the clustering uncertainty by producing multiple plausible clustering configurations. Figure~\ref{fig:spikesorting_smallspikes} shows examples where NCP separates spike clusters with amplitude as low as 3-4$\times$ the standard deviation of the noise into plausible units that are not mere scaled version of each other but have distinct shapes on different channels. f

\section{Conclusion}
We introduced neural architectures
to amortize posterior sampling
of generative clustering models
in $O(N)$ and $O(K)$ forward passes.
The performance is excellent in simple examples. In a realistic spike-sorting application, our results show that NCP spike sorting provides high clustering quality, matches or outperforms a state-of-the-art method, and handles clustering uncertainty 
by efficient posterior sampling (a task that is not solved by currently available methods), demonstrating substantial promise for incorporating these methods into production-scale pipelines.


\clearpage
\newpage

\section*{Acknowledgements}
We thank Sean Bittner, 
Alessandro Ingrosso, Scott Linderman, Aaron Schein and Ruoxi Sun for helpful conversations.
This work was supported by the Simons Foundation, the DARPA NESD program, ONR N00014-17-1-2843, NIH/NIBIB R01 EB22913,
NSF NeuroNex Award DBI-1707398 and The Gatsby Charitable Foundation.

\bibliographystyle{icml2020}
\bibliography{thebib}

\clearpage

\newpage
\appendix
\onecolumn
\setcounter{figure}{0}    
\renewcommand\thefigure{S\arabic{figure}}

\section{Details of the CCP Model}

\subsection{Encodings}
In order to parametrize the  prior, likelihood and posterior of the CCP model, it is convenient to define first 
some symmetric encodings 
for different subsets of the data set $\x$ at iteration $k$. 
Remember that the notation $\x_k$ indicates that 
the dataset is split into three groups, $\x_k = (\x_{\bfa},x_{d_k},\x_{\s})$, where

\begin{center}
\begin{tabular}{ll}
    $\x_{\bfa}$= $ (x_{a_1} \ldots x_{a_{m_k}})$ 
    & $m_k$ available points for  cluster k 
    \\
    $x_{d_k}$&  First data point  in cluster k
    \\
    $\x_{\s}=(\x_{\s_1} \ldots \x_{\s_{k-1}} )$&  
    Points already assigned to clusters. 
\end{tabular}
\end{center}
The symmetric encodings we need are:

\everymath{\displaystyle}

\begin{table}[h!]
\begin{center}
\begin{equation}
\begin{array}{|rl|l|}
\hline 
& \text{Definition}  & \text{Encoded Points} 
\\ 
\hline 
D_k & = 
\sum_{i=1}^{N_k} \delta_{s_{k,i},d_k} u(x_{s_{k,i}})
&
x_{d_k}, \text{ the first point in cluster k}
\\
 A_k^{in} & = \sum_{i=1}^{N_k} (1-\delta_{s_{k,i},d_k}) u(x_{s_{k,i}})
&
\text{Points from $\x_{\bfa}$ that join cluster $k$.}
\\
A_k^{out} & = \sum_{i=1, b_i=0}^{m_k} u(x_{a_i})
& 
\text{Points from $\x_{\bfa}$ that do not join
cluster $k$}
\\
& &
\\
A_k & = A_k^{in}
+ A_k^{out}
&
 \x_{\bfa}, \text{all the $m_k$ points available to join } x_{d_k} 
 \\ & &
\\
 S_k &= D_k + A_k^{in}
&
\text{All points $\s_k$ in cluster k}
\\
\hline 
H_j & = \sum_{x: \, x \in \s_{j}} h(x) \quad  j=1 \ldots k-1  
& \text{All points in cluster $j<k$}
\\
G_k & = \sum_{j=1}^{k-1} g(H_{j})   
&
\text{All the clusters } \s_{1:k-1}.
\\
\hline 
\end{array}
\label{table:encoding}
\end{equation}
\end{center}
\end{table}

\subsection{Prior and Likelihood}
Remember from Section 4 that, having generated $k-1$ clusters $\s_{1:k-1}$,
the  elements of $\s_k$ are generated in a process with latent variables $d_k,\z_k$ 
and joint distribution 
\eqan 
\quad & p_{\theta}(\s_k,\z_k,d_k|\s_{1:k-1},\x)=
p_{\theta}(\bb_k|\z_k, \x_k) p_{\theta}(\z_k|\x_k)p(d_k|\s_{1:k-1})
\enan 
where 
\eqan 
p_{\theta}(\bb_k|\z_k, \x_k)= \prod_{i=1}^{m_k}
p_{\theta,i}(b_{i}|\z_k, \x_k)
\,.
\enan

The priors and likelihood are
\eqan
p(d_k|\s_{1:k-1}) &=& \left\{
\begin{tabular}{cc}
    $1/ |I_k|$ & \textrm{for} $d_k \in I_k$ ,
    \\
    0 &  \textrm{for} $d_k \notin I_k$  ,
\end{tabular}
\right.
\\
p_{\theta}(\z_k|\x_k) &=& 
{\cal N}(\z_k|\mu(\x_k) , \sigma(\x_k))
\label{eq:p1}
\\
p_{\theta,i}(b_{i}|\z_k,\x_k) 
&=& \textrm{sigmoid}[ \rho_i(\z_k, \x_k)] 
\label{eq:p2}
\enan
and can be defined in terms of 
\eqan 
\mu(\x_k)&=&\mu(D_k, A_k, G_k)
\\
\sigma(\x_k)&=&\sigma(D_k, A_k, G_k), 
\\
 \rho_i(\z_k, \x_k) &=&  \rho(\z_k, x_{a_i}, D_k, A_k, G_k)
 \qquad i=1 \ldots m_k
\enan 
where $\mu, \sigma,\rho$
are represented with MLPs. 
Note that in all the cases the functions 
depend on encodings in (\ref{table:encoding}) that are consistent with the permutation symmetries dictated by the conditioning information.

\subsection{ELBO}
The ELBO that we want to maximize is given by 
\eqan
&& 
\mathbb{E}_{p(\x,\s_{1:K})}  \log p_{\theta}(\s_{1:K}|\x) 
\\
&& = 
\mathbb{E}_{p(\x,\s_{1:K})}
\sum_{k=1}^K 
\log\left[\sum_{d_k=1}^{N_k}  \! \! \int \! \! d\z_k  p_{\theta}(\s_k,\z_k,d_k|\s_{1:k-1},\x)
\right]
\\
&& 
\geq 
\mathbb{E}_{p(\x,\s_{1:K})} 
\sum_{k=1}^K 
\mathbb{E}_{q_{\phi}(\z_k,d_k|\s_{1:k},\x)}
\log \left[
\frac{p_{\theta}(\s_k,\z_k,d_k|\s_{1:k-1},\x)}
  {q_{\phi}(\z_k,d_k|\s_{1:k},\x)}
 \right]
\\
&& 
=
\mathbb{E}_{p(\x,\s_{1:K})} 
\sum_{k=1}^K 
\mathbb{E}_{q_{\phi}(\z_k,d_k|\s_{1:k},\x)}
\log \left[
\frac{
p_{\theta}(\bb_k|\z_k, \x_k) p_{\theta}(\z_k|\x_k)p(d_k|\s_{1:k-1})
}  
{
q_{\phi}(\z_k|\bb_k, d_k, \x_k)q_{\phi}(d_k|\s_{1:k}, \x)
}
 \right]
\enan 
where we introduced 
the posterior 
$q_{\phi}(\z_k,d_k|\s_{1:k}, \x) =q_{\phi}(\z_k|\bb_k, d_k, \x_k)
q_{\phi}(d_k|\s_{1:k}, \x)$.
For the first factor we assume a form
\eqan 
q_{\phi}(\z_k|\bb_k, d_k, \x_k) &=&
{\cal N}(\z_k|\mu_q(D_k, A_k^{in}, A_k^{out},G_k), 
\sigma_q(D_k, A_k^{in}, A_k^{out},G_k))
\label{eq:q}
\enan 
where $\mu_q,\sigma_q$ are MLPs.
The most challenging  aspect of maximizing the ELBO
concerns the factor $q_{\phi}(d_k|\s_{1:k}, \x)$,
a multinomial over the $N_k$ components 
of $s_k$  for which we consider next two different approaches.\footnote{
A third alternative would 
be to model $q_{\phi}(d_k| \cdot)$ as (\ref{eq:q_dk}) and compute the
expectation exactly
\eqan 
\mathbb{E}_{p(\x,\s_{1:K})} 
\sum_{k=1}^K 
\sum_{i=1}^{N_k}q_{\phi}(d_k=s_{k,i}|\s_{1:k},\x)
\mathbb{E}_{q_{\phi}(\z_k|\bb_k,d_k,\x_k)}
\log \left[
\frac{
p_{\theta}(\bb_k|\z_k, \x_k) p_{\theta}(\z_k|\x_k)p(d_k|\s_{1:k-1})
}  
{
q_{\phi}(\z_k|\bb_k, d_k, \x_k)q_{\phi}(d_k|\s_{1:k}, \x)
}
 \right]
\enan 
}

\subsection{Gumbel-Softmax Relaxation}
We start by modeling 
\eqan 
q_{\phi}(d_k = s_{k,i}|\s_{1:k}, \x) &=& \textrm{Softmax}[\varphi(x_{s_{k,i}},S_k,A_k^{out},G_k)] \qquad i=1\ldots N_k
\label{eq:q_dk}
\enan 
Following~\citep{jang2016categorical,maddison2016concrete}, we define
\eqan
y_i = \frac{e^{(\varphi_i+g_i)/\tau} }{\sum_{j=1}^{N_k}e^{(\varphi_j+g_j)/\tau}}
\enan 
where $\tau$ is a temperature 
parameter, $g_i$'s are samples from the Gumbel distribution and
\eqan 
\varphi_i = \varphi(x_{s_{k,i}},S_k,A_k^{out},G_k) \,.
\enan 
The  $y_i$'s 
samples are a relaxed version of
one-hot samples of $d_k$ from (\ref{eq:q_dk})
that live in the $N_k$-simplex. 
To apply the relaxation of  $d_k$, 
we just replace 
$\delta_{s_{k,i},d_k}$ with $y_i$
in the definitions of $D_k$ and $A_k^{in}$.
Following the recommendation of~\cite{maddison2016concrete},
we express the ELBO in terms of 
\eqan 
t_i = \log(y)
\enan 
Calling $\kappa_{g,\tau}(t)$ their probability density (see ~\cite{maddison2016concrete} for the explicit form), the relaxed ELBO becomes
\eqan 
\mathbb{E}_{p(\x,\s_{1:K})} 
\sum_{k=1}^K 
\mathbb{E}_{q_{\phi}(\z_k,\y|\s_{1:k},\x)}
\log \left[
\frac{ 
\prod_{i=1}^{N_k} [p_{\theta,i}(b_i=1|\z_k,\y)]^{1-y_i}
\prod_{i=1, b_i=0}^{m_k} p_{\theta,i}(b_i=0|\z_k,\y)
p_{\theta}(\z_k|\y) 
}
{
q_{\phi}(\z_k|\y)
 \kappa_{g,\tau}(t(y))
}
\right] + const.
\nn
\enan 
In this relaxed version the reparametrization trick 
can be used in the usual way to compute derivatives of $q_{\phi}$.

\subsection{Uniform Discrete Posterior}
A simpler approach to model 
$q_{\phi}(d_k|\s_{1:k}, \x)$ is by approximating it as uniform, given by
\eqan
q(d_k|\s_{1:k}) &=& \left\{
\begin{tabular}{cc}
    $1/N_k$ & \textrm{for} $d_k \in \s_k$,
    \\
    0 &  \textrm{for} $d_k \notin \s_k$.
\end{tabular}
\right.
\enan 
This approximation is very good in
cases of well separated clusters.
Since $q(d_k|\s_{1:k})$ has  no parameters now,
this avoids the problem of backpropagation through discrete variables. 
In the examples we considered, this simpler approach yielded better results, as measured, e.g., by a better agreement in Geweke's test (see Figure 5). 
So this was the approach we adopted in the results we present in this work.

\begin{figure*}[t!]
	\begin{center}
		\fbox{\includegraphics[width=.99\textwidth]{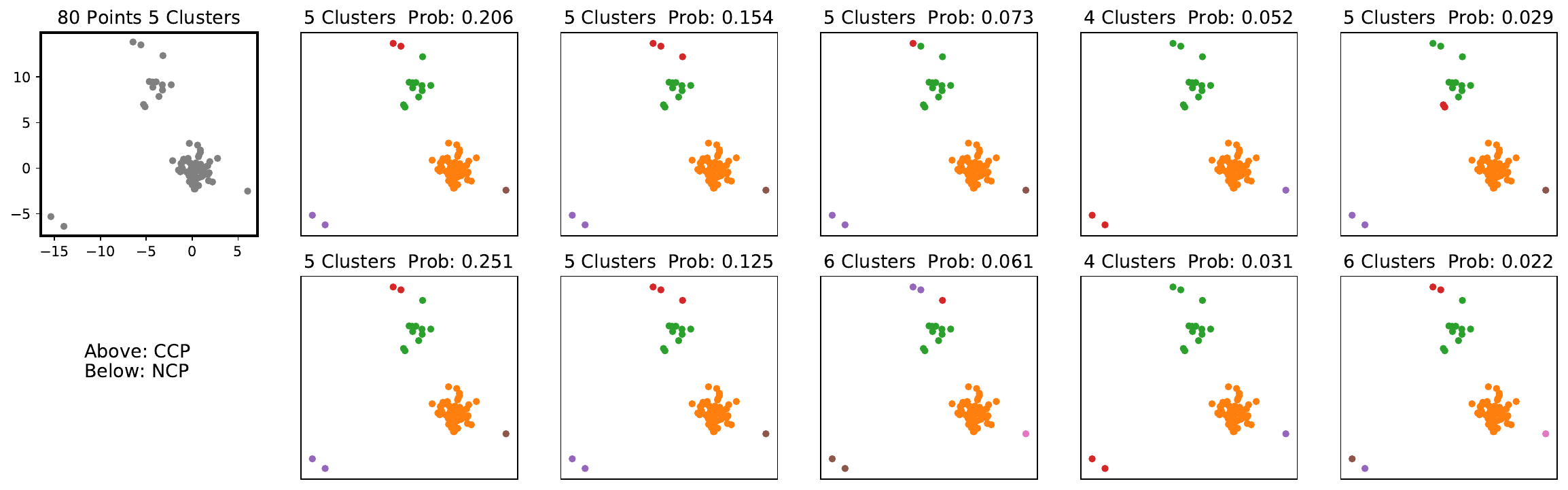}}
		\vskip .2cm
		\fbox
		{\includegraphics[width=.99\textwidth]{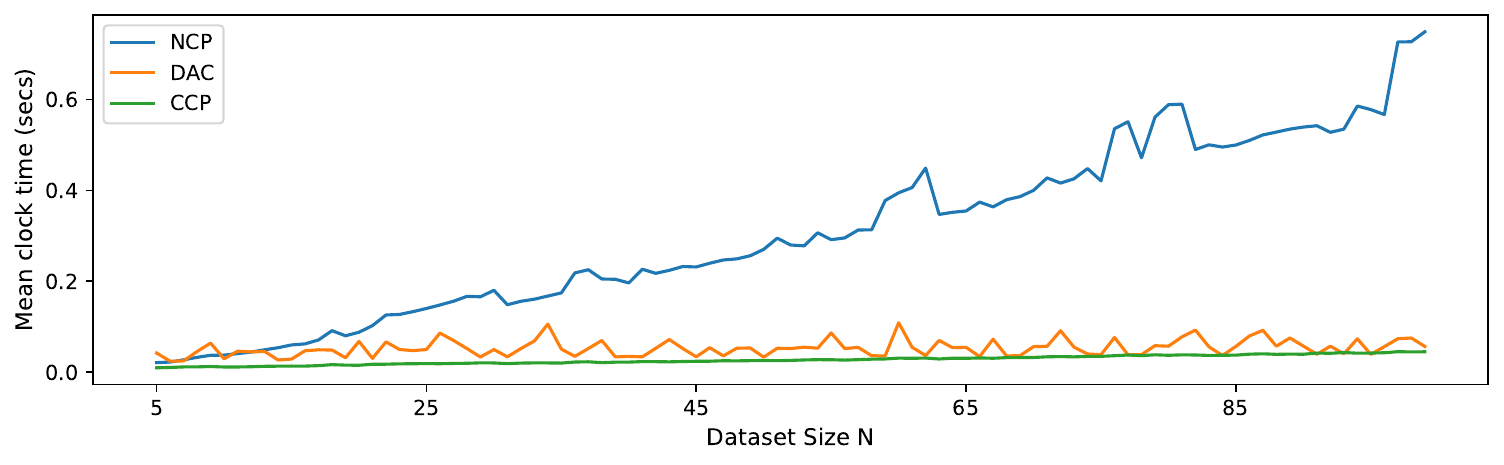}}
	\end{center}
	\vspace{-2mm}
	\caption{
		{
			{\bf Comparing Samples and Times.}
			{\it Above:} 
			Samples from NCP and CCP on the same data set. Both models were trained using the
generative model for mixtures of 2D Gaussians from Section 5.
			Note that the higher probability clusters agree in their labels and approximately in the assigned probabilities. {\it Below:} 
			Clock time as a function of the dataset size
			for NCP, CCP and DAC~\cite{DAC},\footnotemark
			all trained and tested with the same 
			2D Gaussian model as above.
			Each point in the curve is the average over 25 datasets. For NCP and CCP 
			we sampled 200 full posterior samples,
			while DAC gives a single deterministic output. 
		}
	}
	\label{fig:ncp_vs_CCP}
\end{figure*}
\footnotetext{We used the DAC code available at https://github.com/ICLR2020anonymous/dac.}

\newpage
\section{Neural Clustering Process
for Exponential Families}
\label{app:exponential}
\begin{figure*}[t!]
	\begin{center}
		\includegraphics[width=.8\textwidth]{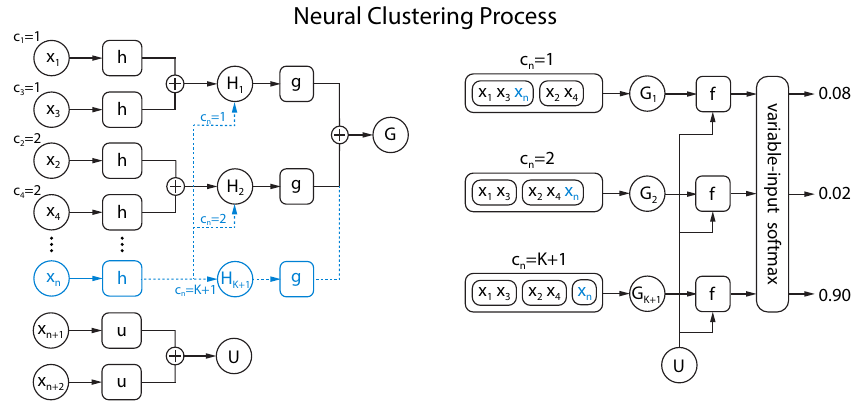}		
	\end{center}
	\vspace{-2mm}
	\caption{
		{
			{\bf Architecture of the Neural Clustering Process.} 
			The full model is composed by the deep networks
			$h,g,u,f$. 
			{\it Left:}
			After assigning the cluster labels $c_{1:n-1}$,
			each possible discrete value $k$  for~$c_n$
			gives a different
			symmetry-invariant encoding 
			of $x_{1:n}$ into 
			the vector
			$G_k$, using the functions $h$ and~$g$.
			The remaining, yet-unassigned
			points $x_{n+1:N}$
			are encoded by $u$ and summed into 
			the vector~$U$.
			{\it Right:}~Each pair $G_k,U$ 
			is mapped by $f$ into a real number (logit), which in turn is 
			mapped into the multinomial 
			distribution  $q_{\theta}(c_n|c_{1:n-1}, \mathbf{ x})$ via a variable-input softmax.
		}
	}
	\label{fig:diagram_ncp}
\end{figure*}

The details of the NCP architecture are fully explained in Section 3,
and Figure~\ref{fig:diagram_ncp} shows the architecture diagramatically.

In the section we consider the spacial case of exponential 
family likelihoods, given by 
\eqan 
p(x|\mu) &=& e^{\mu \cdot t(x)  -\psi(\mu)} m(x)
\\
&=& e^{\lambda \cdot h(x)} m(x)
\enan
where $t(x)$ is a vector of sufficient statistics, and we  defined
\eqan
h(x) &=& (1,t(x))
\\
\lambda &=& (-\psi(\mu),\mu) 
\enan 

Let us denote by $K$ and $K' \geq K$ the total number of distinct values in $c_{1:n}$ and $c_{1:N}$, respectively. 
Consider the joint distribution 
\eqan
p(c_{1:N}, \x, \mu) =  p(c_{1:N}) p(\mu)  \prod_{k=1}^{K'} e^{\lambda_k \cdot\sum_{i:c_i=k}h(x_i)}
\prod_{i=1}^N m(x_i) \,
\enan
from which we obtain the marginal distributions
\eqan
p(c_{1:n}, \x)
&=&
\sum_{c_{n+1}\ldots c_{N}}p(c_{1:N}, \x) 
\\
&=& \sum_{c_{n+1}\ldots c_{N}} \int d\mu  p(c_{1:N}) p(\mu)
\prod_{k=1}^{K'} e^{\lambda_k \cdot
(H_k + \sum_{i>n:c_i=k}h(x_i))}
\prod_{i=1}^N m(x_i)
\\
&=& F(H_1,\ldots, H_K, h(x_{n+1}),\ldots, h(x_{N})  ) \prod_{i=1}^N m(x_i)
\label{def_F}
\enan
where we defined
\eqan 
H_{k} = \sum_{i\leq n, c_i=k} h(x_i)
\qquad k=1\ldots K
\enan 
and $H_k = 0$ for  $k> K$. 

Note now that if $p(c_{1:N})$ is constant, all the dependence of $F$ on $c_{1:n}, x_{1:n}$ is encoded in 
the $H_{k}$'s, and $F$ is symmetric under 
separate permutations of the $H_k$'s and the $h(x_i)$'s for 
$i> n$. 
Based on these symmetries 
we can approximate  $F$ as
\eqan 
F \simeq e^{f(G,U)}
\enan 
modulo adding to $f$ any function symmetric on all
$x_i$'s, 
where 
\eqan 
G &=& \sum_{k=1}^K g(H_k)
\\
U &=& \sum_{i=n+1}^N u(x_i)
\enan 
In the conditional probability we are interested in, 
\eqan
p(c_n|c_{1:n-1}, \x) 
&=&
\frac{p(c_{1:n}, \x)}{\sum_{c_n} p(c_{1:n}, \x)} \,,
\enan 
the product of  the $m(x_i)$'s in (\ref{def_F}) cancels. 
Similarly, 
adding to $f$ 
a function symmetric on all $x_i$'s
leaves invariant our proposed approximation
\eqan 
q_{\theta}(c_n=k|c_{1:n-1}, \mathbf{ x}) = \frac{ e^{f(G_k,U)} }
{  \sum_{k'=1}^{K+1} e^{f(G_{k'},U)}       }  
\qquad k = 1\ldots K+1 \,.
\enan

\section{Monitoring global permutation invariance}
\label{app:monitoring}
\begin{figure}[h!]
	\centering
	\includegraphics[width=.7\textwidth,height=2.6in]{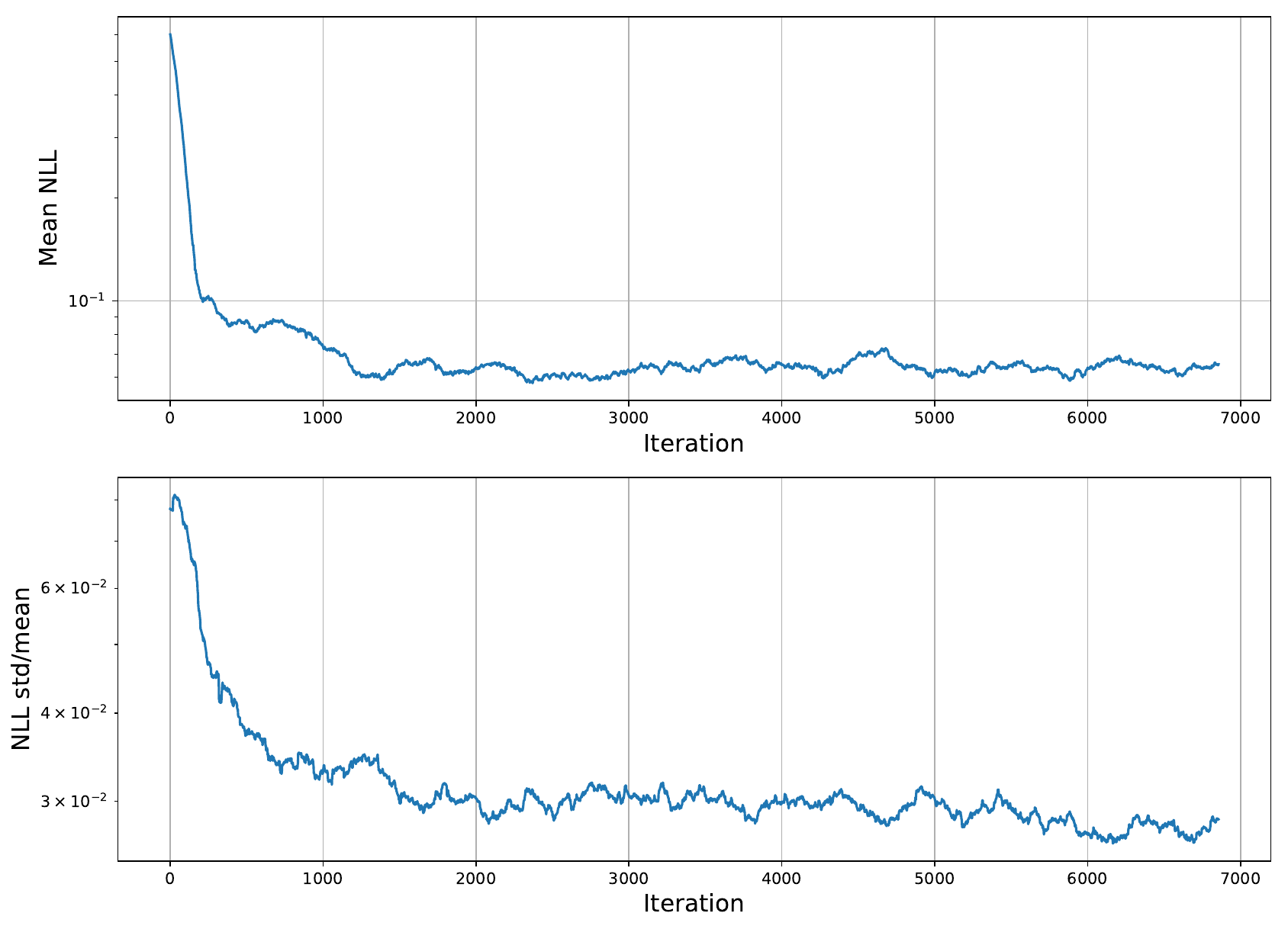}
	\vspace{-2mm}
	\caption{{\bf Global permutation invariance.} 
		Training curves for the NCP model of 2D Gaussians in Section 2. Each minibatch 
		was evaluated for 8 random permutations of the order 
		of the points in the dataset. 
		{\it Above:} Mean of the NLL over the permutations.
		{\it Below:} 
		NLL standard deviation/NLL mean.
		Note that the ratio
		is of order $10^{-2}$.
	}	
	\label{fig:learning}
\end{figure}

As mentioned in Section 7,
we must verify the symmetry of the posterior likelihood 
under global permutations of all the data points.
We show such a check in~\cref{fig:learning}.

\newpage 

\section{Details of spike sorting using NCP}
\label{sec:spike_sorting_methods}

\textbf{Data preprocessing.} 
Training and test data come from the retinal recordings in \cite{Chichilnisky2002} using a 512-channel 2D hexagonal MEA with 20 kHz sampling rate.
After spike detection \citep{lee2017yass}, each multi-channel spike waveform was assigned to the channel where the waveform has the maximum peak-to-peak (PTP) amplitude (i.e. the center channel, ch0). This partitioned the recording data by channel such that each center-channel-based partition only contains multi-channel spike waveforms centered at that channel. Each spike waveform is represented as a 7 $\times$ 32 array containing the 32 time steps surrounding the peak from the center channel and the same time window from the 6 immediate neighbor channels (Figure \ref{fig:spike_sorting}). 
These $7 \times 32$ arrays are the spikes on which clustering was performed.

\textbf{Neural architecture for NCP spike sorting.}
The overall architecture is the same as the one described in Section 3 and Figure \ref{fig:diagram_ncp}. To extract useful features from the spatial-temporal patterns of spike waveforms, we use a 1D ConvNet as the $h$ and $u$ encoder functions. The convolution is applied along the time axis, with each electrode channel treated as a feature dimension. The ConvNet uses a ResNet architecture \citep{He2016DeepResidual} with 4 residual blocks, each having 32, 64, 128, 256 feature maps (kernel size = 3, stride = [1, 2, 2, 2]). The last block is followed by an averaged pooling layer and a final linear layer. The outputs of the ResNet encoder are the $h_i$ and $u_i$
vectors of NCP, i.e. $h_i = \textrm{ResNetEncoder}(x_i)$. We used $d_h=d_u=256$. 
The other two functions, $g$ and $f$, are identical to those in the 2D Gaussian example. 

\textbf{Training NCP using synthetic data.}
To train NCP for spike clustering, we created synthetic labeled training data using a MFM generative model \citep{miller2018mixture} of noisy spike waveforms that mimic the distribution of real spikes:

\vspace{-2ex}
\begin{minipage}{.45\linewidth}
\begin{align}
    N & \sim \textrm{Uniform}[N_{min},N_{max}] \\
    K & \sim 1+ \textrm{Poisson}(\lambda) \\
    \pi_1 \ldots \pi_K & \sim \textrm{Dirichlet} (\alpha_1, \ldots, \alpha_K)
\end{align}
\end{minipage}%
\begin{minipage}{.55\linewidth}
\begin{align}
    c_1 \ldots c_N & \sim \textrm{Cat} (\pi_1, \ldots, \pi_K) \\
    \mu_k & \sim p(\mu) 
    \quad k = 1 \ldots K
    \label{mu_sample}
    \\
    x_i & \sim p(x_i | \mu_{c_i}, \Sigma_s \otimes \Sigma_t) \quad i = 1 \ldots N 
    \label{x_sample}
\end{align}
\end{minipage}

\begin{figure}[b!]
	\centering
	\includegraphics[width=\textwidth]{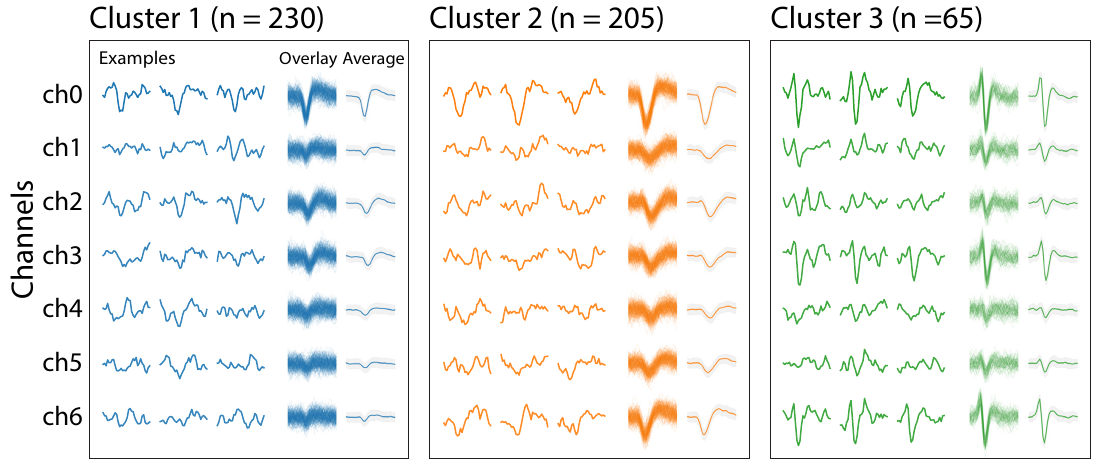}
	\vspace{-2mm}
	\caption{{\bf Synthetic data examples.} Example of 500 synthetic spikes from 3 clusters.  
	}
	\label{fig:spikesorting_synthetic_examples}
\end{figure}

Here, $N$ is the number of spikes between $[200, 500]$. The number of clusters $K$ is sampled from a shifted Poisson distribution with $\lambda = 2$ so that each channel has on average 3 clusters.
$\pi_{1:K}$ represents the proportion of each cluster and is sampled from a Dirichlet distribution with $\alpha_{1:K}=1$.  
The training spike templates $\mu_k \in \mathbb{R}^{7\times 32}$ are sampled  from a reservoir of 957 ground-truth templates not present in any test data, with the temporal axis slightly jittered by random resampling. 
Finally, each waveform $x_i$ is obtained by adding to $\mu_{c_i}$ Gaussian noise with covariance given by the Kronecker product of spatial and temporal correlation matrices estimated from the training data. This method creates spatially and temporally correlated noise patterns similar to real data (Figure \ref{fig:spikesorting_synthetic_examples}). 
We trained NCP for 20000 iterations on a GPU with a batch size of 32 to optimize the NLL loss by the Adam optimizer \citep{kingma2014adam}. 
A learning rate of 0.0001 was used (reduced by half at 10k and 17k iterations).

\textbf{Probabilistic spike clustering using NCP.} At inference time, we fed the  7 x 32 arrays of spike waveforms to NCP, and performed GPU-parallelized posterior sampling of cluster labels (Figure \ref{fig:diagram_ncp} and Figure \ref{fig:spike_sorting}). 
Using beam search \citep{Graves2012SequenceTW, Sutskever2014SequenceTS} with a beam size of 150, we were able to efficiently sample 150 high-likelihood clustering configurations for 2000 spikes in less than 10 seconds on a single GPU. After clustering, we obtained a spike template for each cluster as the average shape of the spike waveforms. The clustering configuration with the highest probability was used for most experiments. 

\textbf{The spike sorting pipelines for real and hybrid data.}
The real data is a 49-channel, 20-minute retina recording with white noise stimulus. To create the hybrid test data, 20 ground-truth spike templates were manually selected from a 49-channel test recording and inserted into another test dataset according to the original spike times. 

For NCP and vGMFM, we performed clustering on 2000 randomly sampled spikes from each channel (clusters containing less than 20 spikes were discarded), and assigned all remaining spikes to a cluster based on the L2 distance to the cluster centers. Then, a final set of unique spike templates were computed,
and each detected spike was assigned to one of the templates. 
The clustering step of vGMFM uses the first 5 PCA components of the spike waveforms as input features.
For Kilosort, we run the entire pipeline using the Kilosort2 package \citep{kilosort2github}. After extracting spike templates and RFs from each pipeline, we matched pairs of templates from different methods by L-infinity distance and pairs of RFs by cosine distance.

\textbf{Electrode drift in real MEA data.} The NCP spike sorting pipeline described above does not take into consideration electrode drift over time, which is present in some real recording data. As a step towards addressing the problem of spike sorting in the presence of electrode drift \citep{calabrese2011kalman,shan2017model}, we describe in Sup. Material \ref{sec:particle_tracking} a generalization of NCP to handle data in which the per-cluster parameters (e.g. the cluster means) are nonstationary in time.

\section{Experimental results for NCP spike sorting}
\label{sec:spike_sorting_results}

\begin{figure}[htb!]
	\centering
	\includegraphics[width=.7\textwidth]{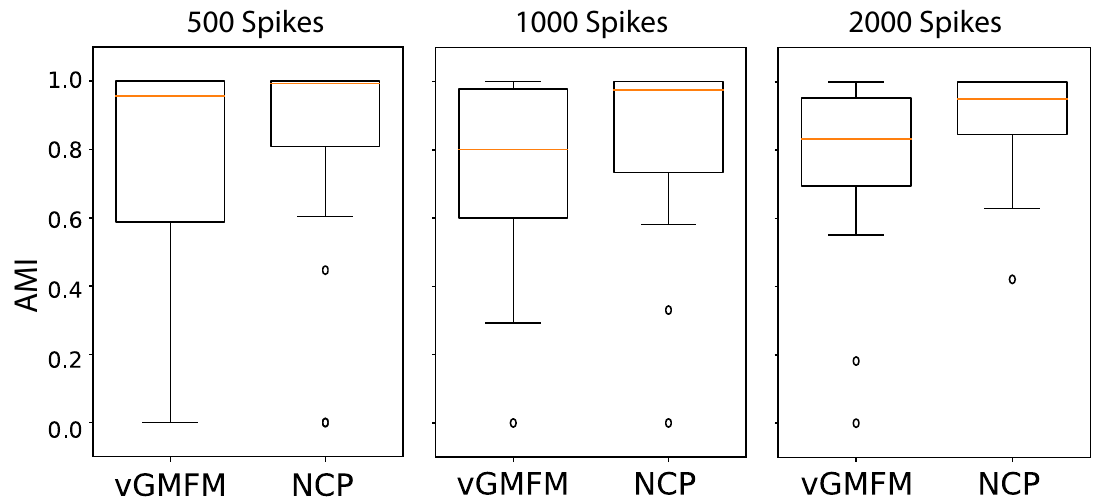}
	\vspace{-2mm}
	\caption{{\bf Clustering synthetic data.} The AMI scores for clustering 20 sets of 500, 1000, and 2000 unseen synthetic spikes. 
	}
	\label{fig:spikesorting_synthetic_results}
\end{figure}

\textbf{Synthetic Data.}
We run NCP and vGMFM on 20 sets of synthetic test data each with 500, 1000, and 2000 spikes.
As the ground-truth cluster labels are known, we compared the clustering quality using Adjusted Mutual Information (AMI) \citep{vinh2010information}. The AMI of NCP is on average 11\% higher than vGMFM (SM Figure~\ref{fig:spikesorting_synthetic_results}), showing better performance of NCP on synthetic data.

\begin{figure}[htb!]
	\centering
	\includegraphics[width=.96\textwidth]{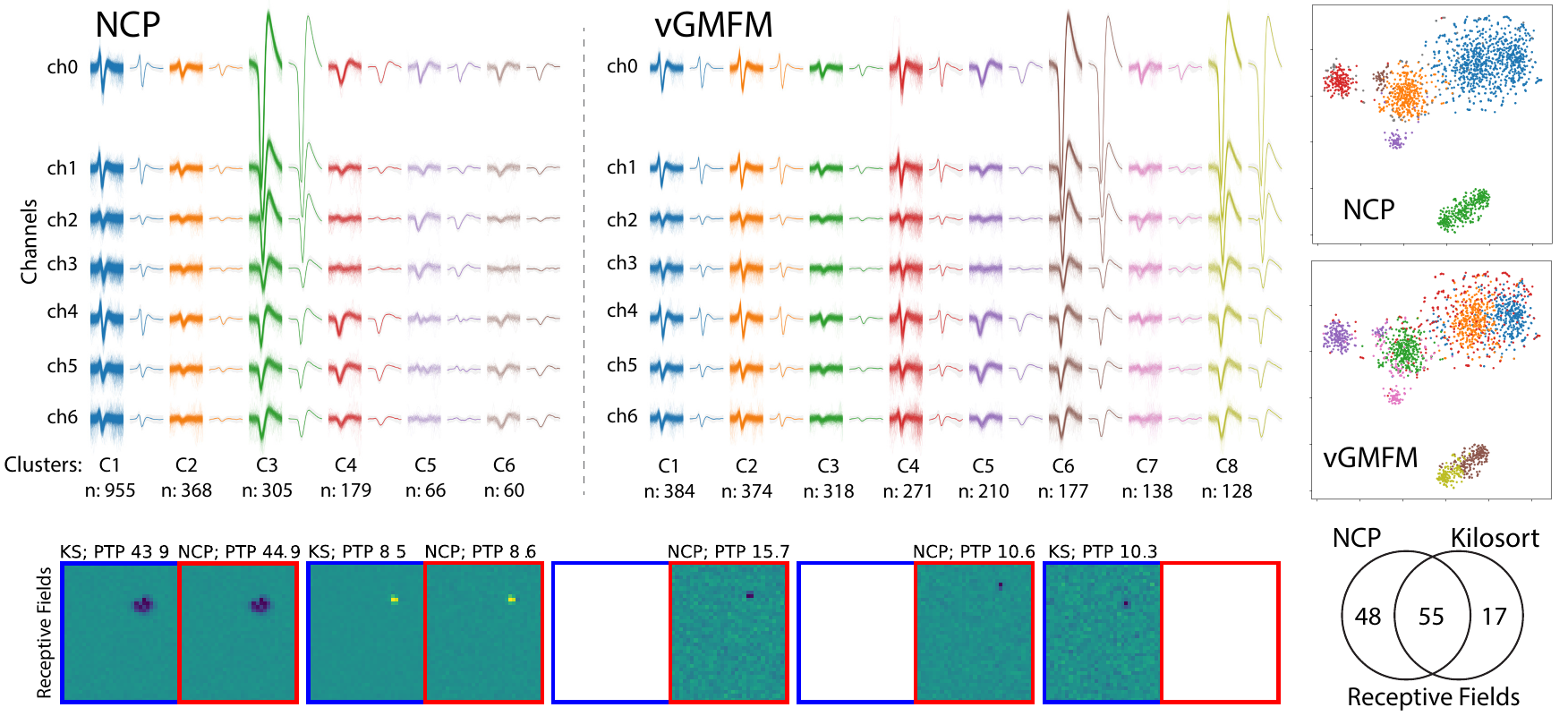}
	\vspace{-2mm}
	\caption{{\bf Spike sorting on real data.} 2000 spikes from real data were clustered by NCP ({\it top-left}) and vGMFM ({\it top-mid}). Each column shows the spikes assigned to one cluster (overlaying traces and their average). Each row is one electrode channel. 
	{\it Top-right:} t-SNE visualization of the spike clusters. 
	{\it Bottom-left:} Example pairs of matched RFs recovered by NCP (red boxes) and Kilosort (blue boxes). Blank indicates no matched counterpart. {\it Bottom-right:} Venn diagram of recovered RFs. 
	}
	\label{fig:spikesorting_real_data}
\end{figure}

\textbf{Real Data.} 
We run NCP, vGMFM and Kilosort on a retina recording with white noise stimulus as described in SM Section~\ref{sec:spike_sorting_methods}, and extracted the averaged spike template of each cluster (i.e. putative neuron). Example 
clustering results in SM Figure~\ref{fig:spikesorting_real_data}
(top) show that NCP produces clean clusters with visually more distinct spike waveforms compared to vGMFM. 
As real data do not come with ground-truth cluster labels, 
we compared the spike templates extracted from NCP and Kilosort using retinal receptive field (RF), which is computed for each cluster as the mean of the stimulus present at each spike. 
A clearly demarcated RF provides encouraging evidence that the spike template corresponds to a real neuron.
Side-by-side comparisons of matched RF pairs are shown in 
SM Figure~\ref{fig:spikesorting_real_data} (bottom-left) 
and SM Figure~\ref{fig:spikesorting_real}. Overall, NCP found 103 templates with clear RFs, among which 48 were not found by Kilosort, while Kilosort found 72 and 17 of them were not found by NCP (SM Figure~\ref{fig:spikesorting_real_data} bottom-right). Thus NCP performs at least as well as Kilosort, and finds many additional templates with clear RFs. 

\begin{figure}[htb!]
    \centering
    \includegraphics[width=.7\textwidth]{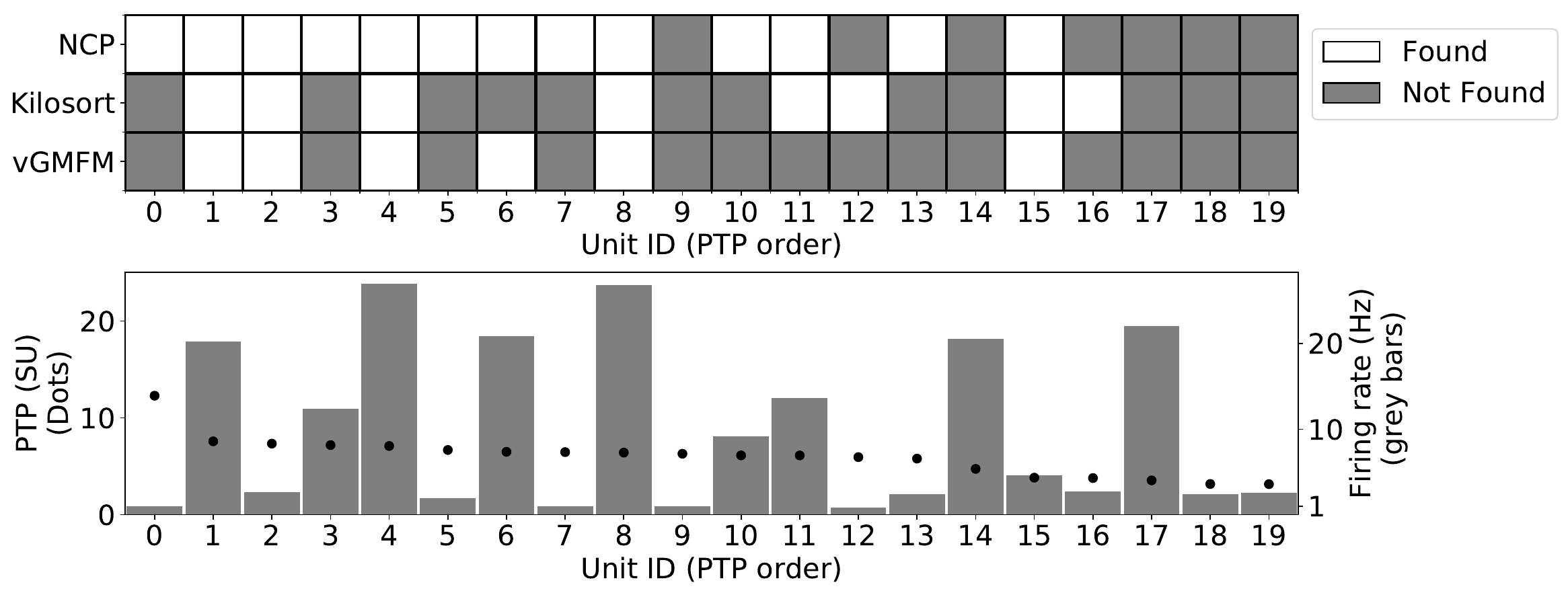}
    \caption{
          {\bf Spike sorting on hybrid data.} {\it Top:} NCP, Kilosort, vGMFM recovered 13, 8, and 6 of the 20 injected ground-truth templates. {\it Bottom:} Peak-to-peak (PTP) size and firing rate of each injected template.  (Smaller templates with lower firing rates are more challenging.)
    	 }
    \label{fig:spikesorting_hybrid}
\end{figure}
\begin{figure}
	\centering
	\includegraphics[width=.8\textwidth]{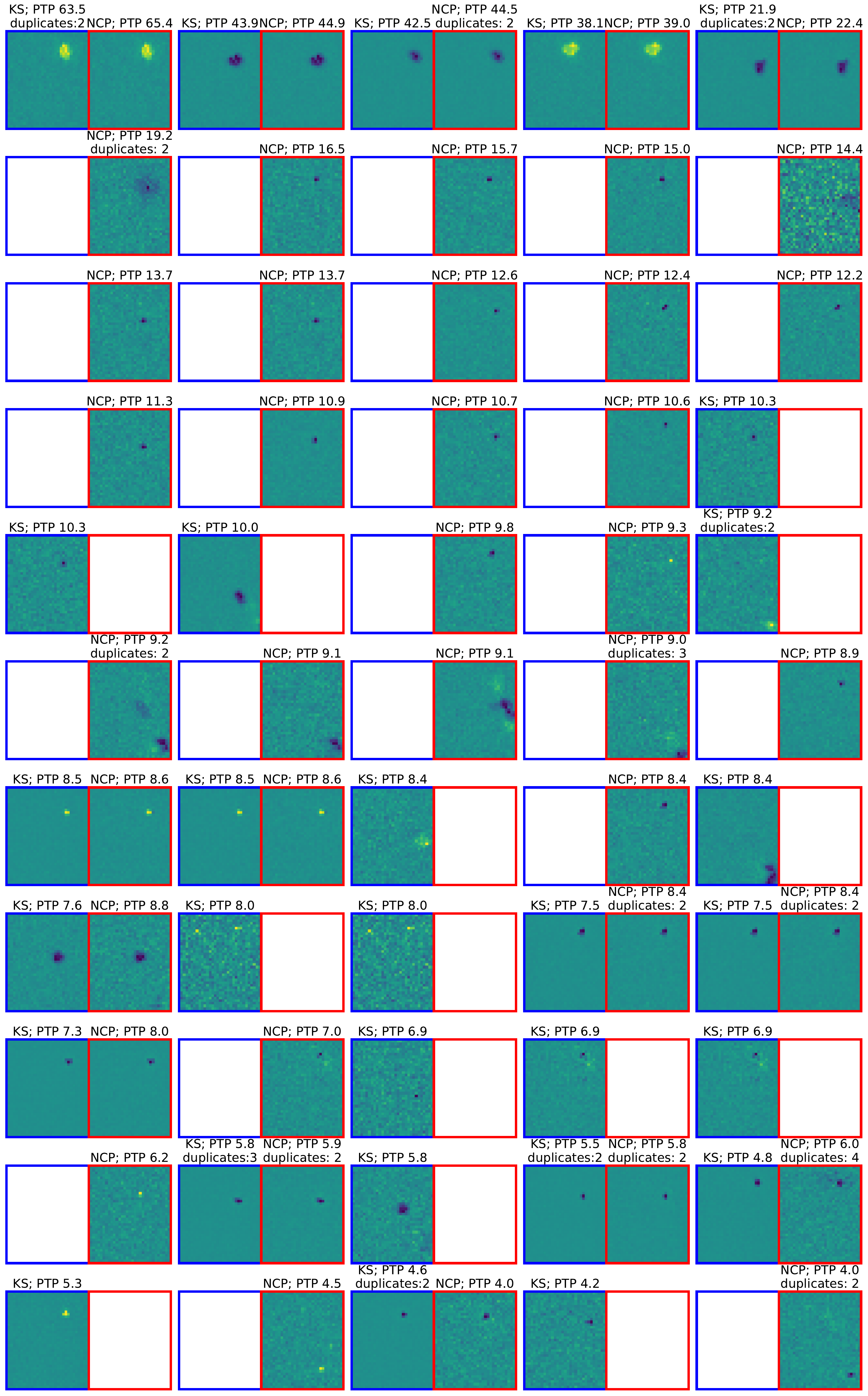}
	\vspace{-2mm}
	\caption{{\bf Spike sorting on real data.} Receptive fields of 55 randomly selected pairs of units recovered from Kilosort and NCP spike sorting.  (Red boxes indicate units found by NCP; blue boxes by Kilosort.) Both approaches find the spikes with the biggest peak-to-peak (PTP) size.  For smaller-PTP units often one sorting method finds a cell that the other sorter misses.  NCP and KS find a comparable number of units with receptive fields here, with NCP finding a few more than KS; see text for details.
	}
	\label{fig:spikesorting_real}
\end{figure}

\textbf{Hybrid Data.}
We compared NCP against vGMFM and Kilosort on a hybrid recording with partial ground truth as in \cite{pachitariu2016kilosort}. Spikes from 20 ground-truth templates were inserted into a real recording 
to test the spike sorting performance on realistic recordings with complex background noise and colliding spikes. 
As shown in SM Figure~\ref{fig:spikesorting_hybrid}, NCP recovered 13 of the 20 injected ground-truth templates, outperforming both Kilosort and vGMFM, which recovered 8 and 6, respectively.


\clearpage

\section{Particle tracking}
\label{sec:particle_tracking}
Inspired by the problem of electrode drift~\citep{calabrese2011kalman,kilosort2github,shan2017model}, 
let us consider now a generative model given by
\eqan
c_t  &\sim& p(c_t|c_{1},\ldots, c_{t-1}) \qquad t=1, \ldots, T
\\
\mu_{k,t} &\sim& p(\mu_{k,t}|\mu_{k,t-1}) \qquad\qquad k=1 \ldots K \qquad t=1, \ldots, T
\label{mu_evo}
\\
x_t &\sim& p(x_t|\mu_{c_{t},t}) \qquad \qquad t=1, \ldots, T
\label{gen4}
\enan
In this model, a cluster 
corresponds to the points along the time trajectory of a particle, and (\ref{mu_evo}) represents the time evolution of the cluster parameters. The cluster labels $c_t$
indicate which particle is observed at time $t$, and
note that particles 
can in principle appear or disappear at any time.

\begin{figure}[t!]
\begin{center}
\fbox{
\includegraphics[width=\textwidth,height=.3\textwidth]{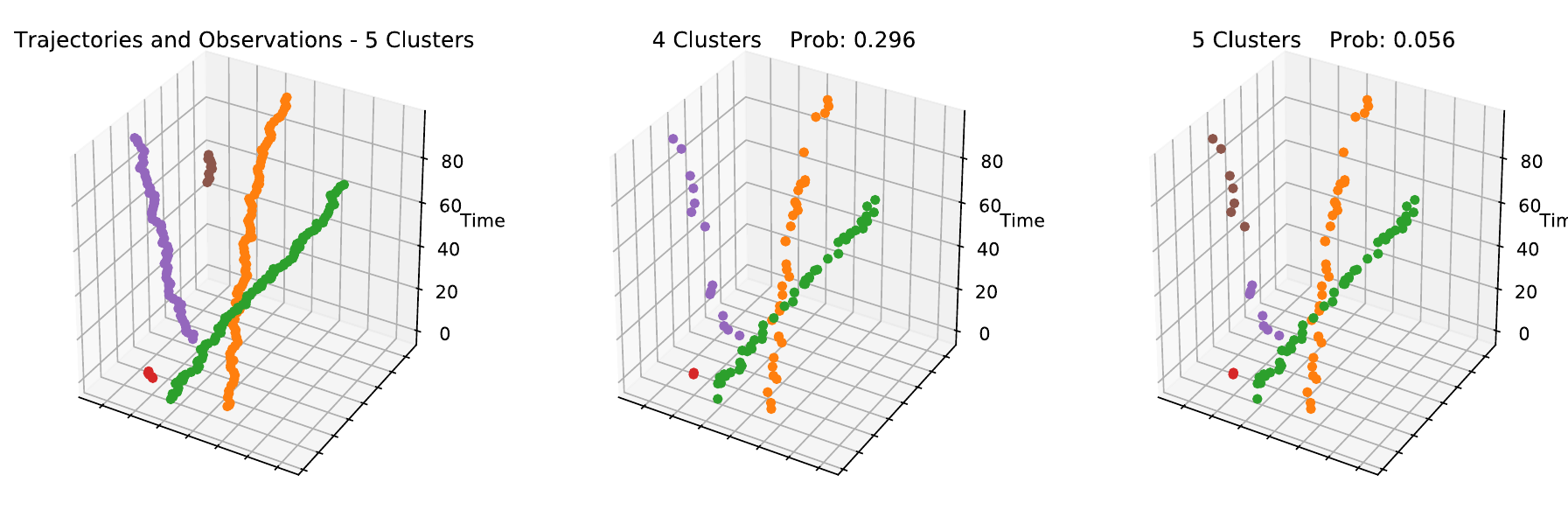}
}
	\end{center}
	\caption{
		{\small 
			{\bf Neural Particle Tracking.} {\it Left:}
			Time trajectories of 5 2D particles. Note that particles can appear or disappear at arbitrary times.
			{\it Middle and right:} Two posterior 
			samples. 
			Note that since only one particle is observed at each time,
			a particle not observed for some time leads to a possible ambiguity on the number of particles.
			 (Best seen in color.)
		}
	}
	\label{fig:ntp}
\end{figure}

To take the time evolution 
into account,
we let particles influence one another with a weight that depends on their time distance. 
For this, let us introduce a  time-decay constant~$b>0$, and generalize the NCP equations to 
\eqan 
H_{k,t} &=& \sum_{t'=1 : c_{t'}=k}^{t} e^{-b|t-t'|} h(x_{t'})  \qquad k = 1\ldots K\,,
\\
G_t &=& \sum_{k =1}^K g(H_{k,t}) \,,
\\
U_t &=& \sum_{t'=t+1}^{T}  e^{-b|t-t'|}  u(x_{t'})  \,.
\enan 		
The conditional assignment probability for $c_t$ is now
\eqan 
q_{\theta}(c_t=k|c_{1:t-1}, \mathbf{ x}) = \frac{ e^{f(G_{k,t},U_t )} }
{  \sum_{k'=1}^{K+1} e^{f(G_{k',t},U_t)}       } 
\label{tncp}
\enan
for $k = 1\ldots K+1$. The time-decay constant $b$ 
is learnt along with all the other parameters. We can also consider
replacing $e^{-b|t-t'|}$ with a general distance function $e^{-d(|t-t'|)}$.
Figure~\ref{fig:ntp} 
illustrates this model in a simple 2D example.
We call this approach Neural Particle Tracking.

\section{Neural architectures in the examples}
\label{app:details}
To train the networks in the examples, we 
used stochastic gradient descent with Adam~\citep{kingma2014adam},
with learning rate $10^{-4}$.
The number of samples in each mini-batch were: 1 for $p(N)$, 
1 for $p(c_{1:N})$, 64 for $p(\x|c_{1:N})$. 
The architecture of the functions
in each case were:

\subsection*{NCP: 2D Gaussians}
\begin{itemize}
    \item $h$: MLP [2-256-256-256-128] with ReLUs
    \item $u$: MLP [2-256-256-256-128] with ReLUs
    \item $g$: MLP [128-256-256-256-256] with ReLUs
    \item $f$: MLP [384-256-256-256-1] with ReLUs
\end{itemize}

\subsection*{NCP: MNIST}
\begin{itemize}
\item $h$: 2 layers of [convolutional + maxpool + ReLU] + MLP [320-256-128] with ReLUs
\item $u$: same as $h$
    \item $g$: MLP [256-128-128-128-128-256] with ReLUs
\item $f$: MLP [384-256-256-256-1] with ReLUs
\end{itemize}


\end{document}